\documentclass[10pt,twocolumn,letterpaper]{article}

\usepackage{titlefoot}
\usepackage{adjustbox}
\usepackage{cvpr}
\usepackage{times}
\usepackage{epsfig}
\usepackage{graphicx}
\usepackage{amsmath}
\usepackage{amssymb}
\usepackage{bbm}
\usepackage{booktabs}
\usepackage{enumitem}
\usepackage{subfig}
\usepackage{authblk}
\usepackage[T1]{fontenc}

\usepackage[breaklinks=true,bookmarks=false]{hyperref}

\cvprfinalcopy 

\def\cvprPaperID{4994} 

\newcommand*\samethanks[1][\value{footnote}]{\footnotemark[#1]}

\makeatletter
\renewcommand\AB@affilsepx{, \protect\Affilfont}
\makeatother

\begin{document}

\title{Argoverse: 3D Tracking and Forecasting with Rich Maps}

\author[1,2]{Ming-Fang Chang\thanks{Equal contribution.} }
\author[3]{John Lambert\samethanks }
\author[1,3]{Patsorn Sangkloy\samethanks }
\author[1]{Jagjeet Singh\samethanks}
\author[ ]{S{\l}awomir B\k{a}k}
\author[1]{Andrew Hartnett}
\author[1]{De Wang}
\author[1]{Peter Carr}
\author[1,2]{Simon Lucey}
\author[1,2]{Deva Ramanan}
\author[1,3]{James Hays}
\affil[1]{Argo AI}
\affil[2]{Carnegie Mellon University}
\affil[3]{Georgia Institute of Technology}



\twocolumn[{%
\renewcommand\twocolumn[1][]{#1}%
\maketitle
\begin{center}
    \centering
    \includegraphics[width=0.8\textwidth]{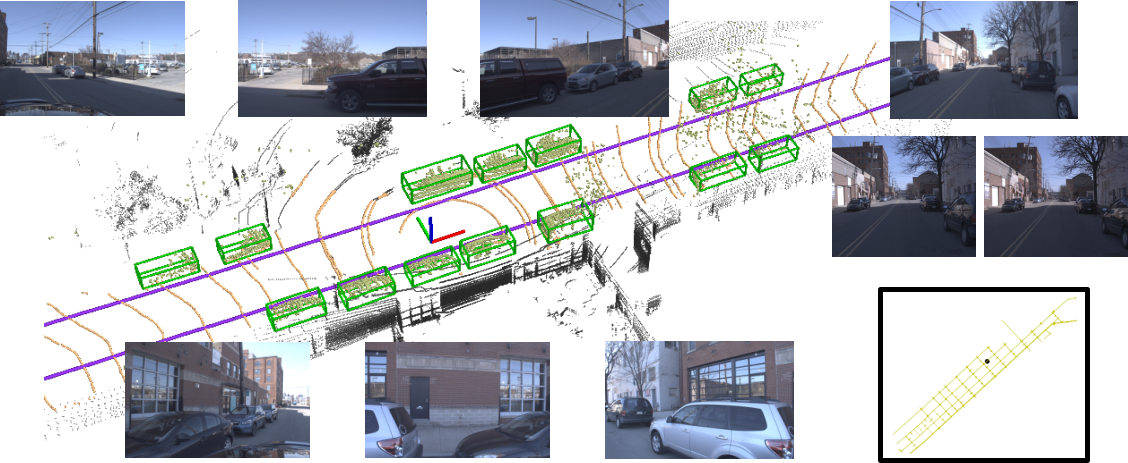}
    \captionof{figure}{We introduce datasets for 3D tracking and motion forecasting with \emph{rich maps} for autonomous driving. Our 3D tracking dataset contains sequences of LiDAR measurements, 360$^{\circ}$ RGB video, front-facing stereo (middle-right), and 6-dof localization. All sequences are aligned with maps containing lane center lines (magenta), driveable region (orange), and ground height. Sequences are annotated with 3D cuboid tracks (green). A wider map view is shown in the bottom-right.}\label{fig:teaser}
\end{center}%
}]



\begin{abstract} \unmarkedfntext{*Equal contribution} We present \emph{Argoverse} -- two datasets designed to support autonomous vehicle machine learning tasks such as 3D tracking and motion forecasting. Argoverse was collected by a fleet of autonomous vehicles in Pittsburgh and Miami. The Argoverse 3D Tracking dataset includes 360$^{\circ}$ images from 7 cameras with overlapping fields of view, 3D point clouds from long range LiDAR, 6-DOF pose, and 3D track annotations. Notably, it is the only modern AV dataset that provides forward-facing stereo imagery. The Argoverse Motion Forecasting dataset includes more than 300,000 5 second tracked scenarios with a particular vehicle identified for trajectory forecasting. Argoverse is the first autonomous vehicle dataset to include ``HD maps'' with 290 km of mapped lanes with geometric and semantic metadata. All data is released under a Creative Commons license at \url{www.argoverse.org}. 
In our baseline experiments, we illustrate how detailed map information such as lane direction, driveable area, and ground height improves the accuracy of 3D object tracking and motion forecasting. 
Our tracking and forecasting experiments represent only an initial exploration of the use of rich maps in robotic perception. We hope that Argoverse will enable the research community to explore these problems in greater depth.



\end{abstract}

\section{Introduction} 

Datasets and benchmarks for a variety of perception tasks in autonomous driving have been hugely influential to the computer vision community over the last few years. We are particularly inspired by the impact of KITTI~\cite{KITTI:geiger2013vision}, which opened and connected a plethora of new research directions. However, publicly available datasets for autonomous driving rarely include \emph{map} data, even though detailed maps are critical to the development of real world autonomous systems. Publicly available maps, \eg OpenStreetMap, can be useful, but have limited detail and accuracy. 

Intuitively, 3D scene understanding would be easier if maps directly told us which 3D points belong to the road, which belong to static buildings, which lane a tracked object is in, how far it is to the next intersection, etc. But since publicly available datasets do not contain richly-mapped attributes, \textit{how} to represent and utilize such features is an open research question. Argoverse is the first large-scale autonomous driving dataset with such detailed maps. We investigate the potential utility of these new map features on two tasks -- 3D tracking and motion forecasting, and we offer a significant amount of real-world, annotated data to enable new benchmarks for these problems.

Our contributions in this paper include:
\begin{itemize}
    \itemsep0em 
    \item We release a large scale 3D tracking dataset with synchronized data from LiDAR, 360$^{\circ}$ and stereo cameras sampled across two cities in varied conditions. Unlike other recent datasets, our 360$^{\circ}$ is captured at 30fps.
    \item We provide ground truth 3D track annotations across 15 object classes, with five times as many tracked objects as the KITTI~\cite{KITTI:geiger2013vision} tracking benchmark.
    \item We create a large-scale forecasting dataset consisting of trajectory data for interesting scenarios such as turns at intersections, high traffic clutter, and lane changes. 
    
    \item We release map data and an API which can be used to develop map-based perception and forecasting algorithms. We are the first self-driving vehicle dataset with a semantic vector map of road infrastructure and traffic rules.  The inclusion of ``HD'' map information also means our dataset is the first large-scale benchmark for automatic map creation, often known as \textit{map automation}.
    \item We are the first to examine the influence of HD map context for 3D tracking and motion forecasting. In the case of 3D tracking, we measure the influence of map-based ground point removal and orientation snapping to lanes. In the case of motion forecasting, we experiment with the creation of diverse predictions from the lane graph and the pruning of predictions by the driveable area map. In both cases, we see higher accuracy with the use of a map.
    
\end{itemize}

\section{Related Work}


\noindent \textbf{Autonomous Driving Datasets with Map Information}.
Until recently, it was rare to find datasets that provide detailed map information associated with annotated data. The prohibitive cost of annotating and constructing such maps has spurred interest in the growing field of map automation \cite{Liang:2019:CVPR:roadboundary, Homayounfar_2018_CVPR, Bai:2018:IROS:deep-ms-lane-det}. Prior to Argoverse's release, no public dataset included 3D vector map information, thus preventing the development of common benchmark for map automation. TorontoCity~\cite{wang2016torontocity} also focuses on map construction tasks but without 3D annotation for dynamic objects.
The nuScenes dataset~\cite{holger:nuscenes:2019} originally contained maps in the form of binary, rasterized, top-down indicators of region of interest (where region of interest is the union of driveable area and sidewalk). This map information is provided for 1000 annotated vehicle log segments (or ``scenes'') in Singapore and Boston. Subsequent to Argoverse release, nuScenes has released labels for 2D semantic map regions, without a lane or graph structure. 
Like nuScenes, we include maps of driveable area, but also include ground height and a ``vector map'' of lane centerlines and their connectivity.


\noindent \textbf{Autonomous Driving Datasets with 3D Track Annotations}.
Many existing datasets for object tracking focus on pedestrian tracking from image/video sequences \cite{Guo:2019:safetodrive,patino2016pets,MOT16,andriluka2008people}. Several datasets provide raw data from self-driving vehicle sensors, but without any object annotations~\cite{maddern20171,Pandey:2011:FordVisionLiDAR, rangesh:LISAdataset:2017}. The ApolloCar3D dataset~\cite{ApolloCar3D:song:cvpr:2019} is oriented towards 3D semantic object keypoint detection instead of tracking. KITTI~\cite{KITTI:geiger2013vision} and H3D~\cite{patil:h3d:ICRA:2019} offer 3D bounding boxes and track annotations but do not provide a map. The camera field of view is frontal, rather than $360^{\circ}$. VIPER \cite{Richter_2017} provides data from a simulated world with 3D track annotations.
nuScenes~\cite{holger:nuscenes:2019} currently provides $360^{\circ}$ data and a benchmark for 3D object detection, with tracking annotation also available.
The \emph{Argoverse 3D Tracking} dataset contains 360$^{\circ}$ track annotations in 3D space aligned with detailed map information. 
See Table~\ref{table:DB_compare} for a comparison between 3D autonomous vehicle datasets.

\noindent \textbf{Autonomous Driving Datasets with  Trajectory Data.} ApolloScape \cite{ApolloScape} also uses sensor-equipped vehicles to observe driving trajectories in the wild and presents a forecasting benchmark \cite{ma:trafficpredict:2019:aaai} from a subset of the ApolloScape 3D tracking annotations. This dataset consists of 155 minutes of observations compared to 320 hours of observations in the \emph{Argoverse Forecasting} dataset. IntentNet \cite{Casas:2018:IntentNet} mines roof-mounted LiDAR data for 54 million object trajectories, but the data is not publicly available.

\noindent \textbf{Using Maps for Self-driving Tasks}.
While high definition (HD) maps are widely used by motion planning systems, few works explore the use of this strong prior in perception systems \cite{HDNET:2018:Yang} despite the fact that the three winning entries of the 2007 DARPA Urban Challenge relied on a DARPA-supplied map -- the \emph{Route Network Definition File} (RNDF) \cite{Montemerlo:2008:JSE:1405647.1405651,Urmson-2007-9708,Bacha:2008:OTV:1395073.1395074}. 
Hecker~\textit{et al.} \cite{ECCV:2018:EndToEnd:Hecker} show that end-to-end route planning can be improved by processing rasterized maps from OpenStreetMap and TomTom.
Liang ~\textit{et al.}~\cite{Liang:2018:ECCV:crosswalks} demonstrate that using road centerlines and intersection polygons from OpenStreetMap can help infer crosswalk location and direction. 
Yang \textit{et al.} \cite{HDNET:2018:Yang} show that incorporating ground height and bird's eye view (BEV) road segmentation with LiDAR point information as a model input can improve 3D object detection. Liang \textit{et al.} \cite{Liang:2019:CVPR:multitask-multisensor} show how 3D object detection accuracy can be improved by using mapping (ground height estimation) as an additional task in multi-task learning.
Suraj \textit{et al.} \cite{suraj:2018:crowdsourcedmaps} use dashboard-mounted monocular cameras on a fleet of vehicles to build a 3D map via city-scale structure-from-motion for localization of ego-vehicles and trajectory extraction. 

\begin{table*}
\center
\begin{adjustbox}{max width=\textwidth}
\begingroup
\renewcommand{\arraystretch}{1} 
 \begin{tabular}{cccccccccc} 
 \textsc{Dataset Name}   & \textsc{Map} & \textsc{Extent of } & \textsc{Driveable } & \textsc{Camera } & \textsc{$360^{\circ}$  }   & \textsc{Includes} & \# \textsc{Tracked} & \# \textsc{Scenes} \\
 & \textsc{Type} & \textsc{Annotated } & \textsc{Area } & \textsc{Frame} & \textsc{Cameras}  & \textsc{Stereo} & \textsc{Objects} & \\ 
 & & \textsc{Lanes} & \textsc{Coverage} & \textsc{Rate} & & & \textsc{/Scene} & \\ 
\midrule
KITTI~\cite{KITTI:geiger2013vision} &  None & 0 km & 0 $\text{m}^2$  & 10 Hz & no   & \checkmark & 43.67 (train)  & 50\\ 
Oxford RobotCar~\cite{maddern20171} &  None & 0 km & 0 $\text{m}^2$  &  11/16Hz & no &   no & 0 & 100+\\ 
H3D \cite{patil:h3d:ICRA:2019} & None & 0 km & 0 $\text{m}^2$ & 30 Hz & no &   no & 86.02 (train+val+test) & 160\\ 
Lyft Dataset \cite{lyft2019} \footnote{on the currently release training set} & Raster & 0 km & 48,690 $\text{m}^2$ & 10 Hz & \checkmark &   no & 102.34 (train) & 180+ \\ 
nuScenes v1.0  \cite{holger:nuscenes:2019} & Vector+Raster & 133 km & 1,115,844 $\text{m}^2$  & 12 Hz & \checkmark &  no & 75.75 (train+val) & 1000\\ 

ApolloScape Tracking \cite{ma:trafficpredict:2019:aaai}& None & 0 km & 0 $\text{m}^2$  & n/a & no &  no & 206.16 (train) & 103\\ 

Waymo Open Dataset 
& None & 0 km & 0 $\text{m}^2$  & 10 Hz & \checkmark &  no & 113.68 (train+val) & 1000\\ 
\midrule
Argoverse 3D Tracking v1.1 & Vector & 204 $\text{km}$ (MIA)   &   1,192,073 $\text{m}^2$ & 30 Hz & \checkmark &   \checkmark & 97.81 (train+val+test) & 113\\ 
(human annotated)  & +Raster & +86 km (PIT) & &  & & & &\\ 
\midrule
ApolloScape Forecasting \cite{ma:trafficpredict:2019:aaai} & None & 0 km & 0 $\text{m}^2$  & n/a & no &  no & 50.06 (train) & 103\\ 
\midrule
Argoverse Forecasting v1.1  & Vector & 204 $\text{km}$ (MIA)   &  1,192,073 $\text{m}^2$  & - & no &  no &  50.03 (train+val+test)  & 324,557\\ 
(mined trajectories) & +Raster & +86 km (PIT) & &  &  & & & &  \\ 
\bottomrule 
\end{tabular}
\endgroup
\end{adjustbox}
\caption{\textbf{Public self-driving datasets}. We compare recent, publicly available self-driving datasets with 3D object annotations for tracking (top) and trajectories for forecasting (bottom). Coverage area for nuScenes is based on its \emph{road and sidewalk} raster map. Argoverse coverage area is based on our \emph{driveable area} raster map. Statistics updated September 2019.
}\label{table:DB_compare}
\end{table*}


\noindent \textbf{3D Object Tracking}. 
In traditional approaches for point cloud tracking, segments of points can be accumulated using clustering algorithms such as DBSCAN \cite{Ester96adensity-based, Leonard:2008:journalfieldrobotics} or connected components of an occupancy grid \cite{Levinson:2011:towardsfully, himmelsbach:08:lidar-based3d}, and then associated based on some distance function using the Hungarian algorithm. Held \textit{et al.} utilize probabilistic approaches to point cloud segmentation and tracking \cite{Held:RSS:2016, Held:RSS:2014, held:2013:precision}.  Recent work demonstrates how 3D instance segmentation and 3D motion (in the form of 3D scene flow, or per-point velocity vectors) can be estimated directly on point cloud input with deep networks \cite{wang:cvpr2018:sgpn, liu:cvpr2019:flownet3d}. Our dataset enables 3D tracking with sensor fusion in a $360^{\circ}$ frame. 

\noindent \textbf{Trajectory Forecasting.} 
Spatial context and social interactions can influence the future path of pedestrians and cars. 
Social-LSTM\cite{socialLSTM} proposes a novel pooling layer to capture social interaction of pedestrians. Social-GAN \cite{Gupta_2018_CVPR} attempts to model the multimodal nature of the predictions. However, both have only been tested on pedestrian trajectories, with no use of static context (e.g. a map). Deo et al. \cite{deo2018convolutional} propose a convolutional social pooling approach wherein they first predict the maneuver and then the trajectory conditioned on that maneuver. 
In the self-driving domain, the use of spatial context is of utmost importance and it can be efficiently leveraged from the maps. Chen et al.~\cite{chen2015deepdriving} use a feature-driven approach for social and spatial context by mapping the input image to a small number affordances of a road/traffic state. However, they limit their experiments to a simulation environment. IntentNet \cite{Casas:2018:IntentNet} extends the joint detection and prediction approach of Luo et al.~\cite{FaF:Luo:2018:CVPR} by discretizing the prediction space and attempting to predict one of eight common driving maneuvers. 
DESIRE \cite{desire_LeeCVCTC17} demonstrates a forecasting model 
capturing both social interaction and spatial context. The authors note that the benefits from these two additional components are small on the KITTI dataset, attributing this to the minimal inter-vehicle interactions in the data. Another challenging problem in the trajectory forecasting domain is to predict diverse trajectories which can address multimodal nature of the problem. R2P2 \cite{rhinehart2018r2p2} address the diversity-precision trade-off of generative forecasting models and formulate a symmetric cross-entropy training objective to address it. It is then followed by PRECOG \cite{rhinehart2019precog} wherein they present the first generative multi-agent forecasting method to condition on agent intent. They achieve state-of-the-art results for forecasting methods in real
(nuScenes \cite{holger:nuscenes:2019}) and simulated (CARLA \cite{dosovitskiy2017carla}) datasets. 


\section{The Argoverse Dataset}
Our sensor data, maps, and annotations are the primary contribution of this work. We also provide an API which connects the map data with sensor information \eg ground point removal, nearest centerline queries, and lane graph connectivity; see the Appendix for more details. The data is available at \url{www.argoverse.org} under a Creative Commons license. The API, tutorials, and code for baseline algorithms are available at \url{github.com/argoai/argoverse-api} under an MIT license. The statistics and experiments in this document are based on Argoverse v1.1 released in October 2019.

We collected raw data from a fleet of autonomous vehicles (AVs) in Pittsburgh, Pennsylvania, and Miami, Florida, both in the USA. These cities have distinct climate, architecture, infrastructure, and behavioral patterns. The captured data spans different seasons, weather conditions, and times of the day. The data used in our dataset traverses nearly 300 km of mapped road lanes and comes from a subset of our fleet operating area.

\begin{figure}
\centering
\includegraphics[width=\linewidth]{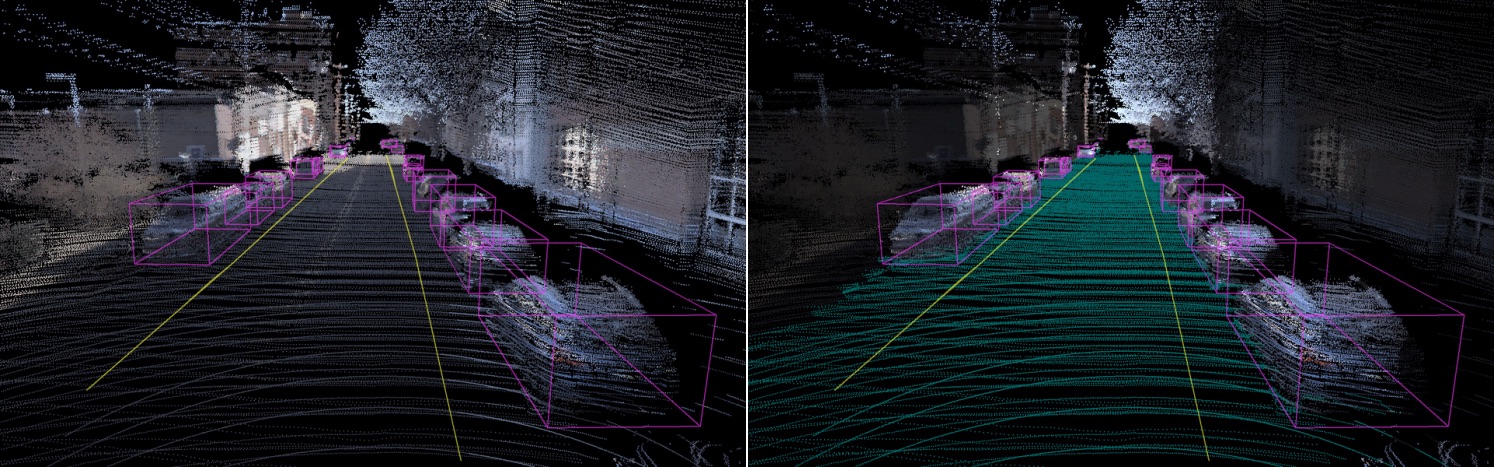}
\caption{\textbf{3D visualization of an Argoverse scene.} Left: we accumulate LiDAR points and project them to a virtual image plane. Right: using our map, LiDAR points beyond driveable area are dimmed and points near the ground are highlighted in cyan. Cuboid object annotations and road centerlines are shown in pink and yellow.}\label{fig:3D_vis}
\end{figure}

\begin{figure}
\centering
\includegraphics[width=0.8\linewidth]{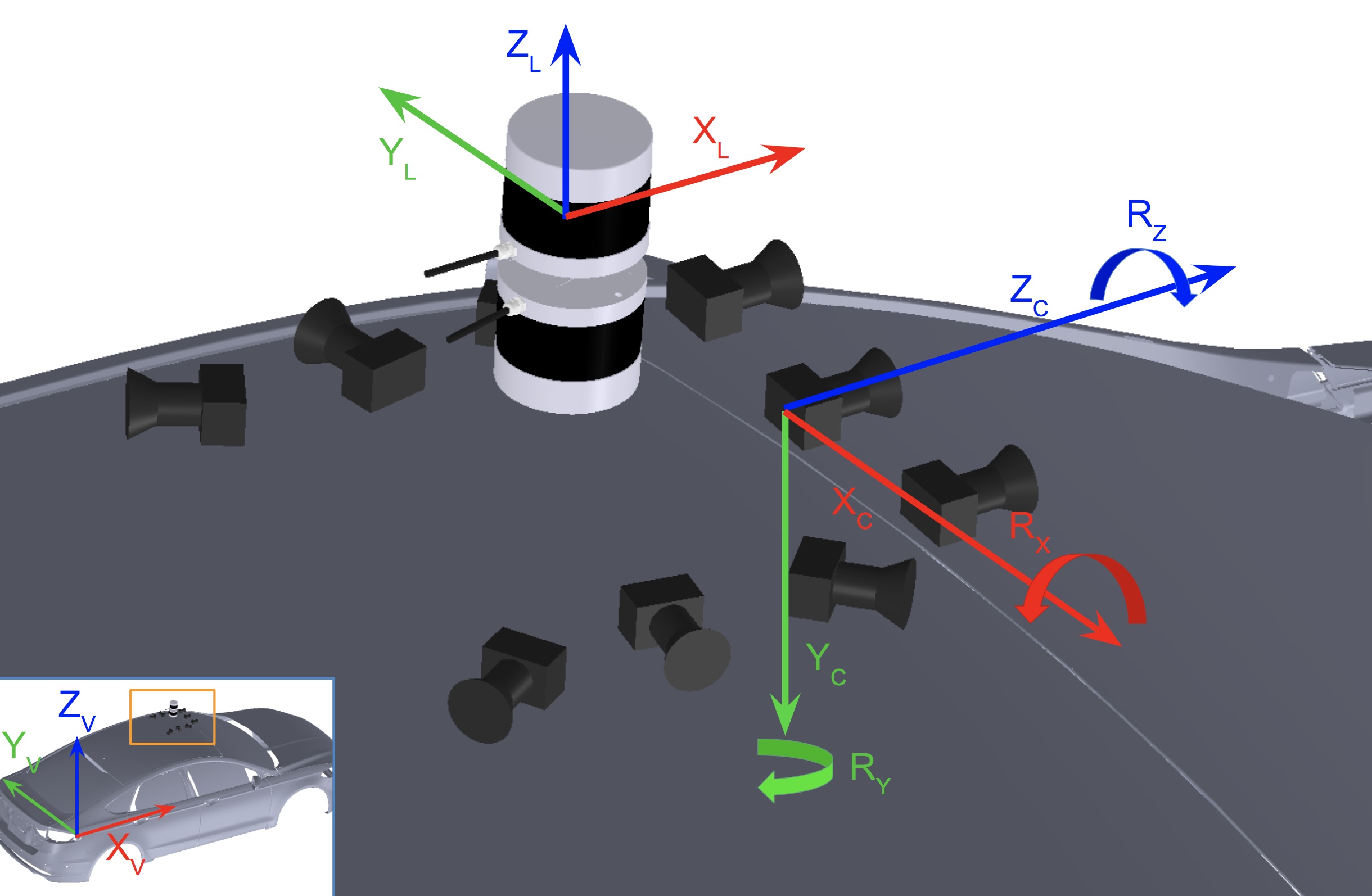}
\caption{\textbf{Car sensor schematic.} Three reference coordinate systems are displayed: (1) the \emph{vehicle frame}, with $X_v$ forward, $Y_v$ left, and $Z_v$ up, (2) the \emph{camera frame}, with $X_c$ across image plane, $Y_c$ down image plane, and $Z_c$ along optical axis, (3) the \emph{LiDAR} frame, with $X_L$ forward, $Y_L$ left, and $Z_L$ up. Positive rotations $R_X, R_Y, R_Z$ are defined for each coordinate system as rotation about the respective axis following the right-hand rule.}\label{fig:sensors}
\end{figure}

\noindent \textbf{Sensors}. Our vehicles are equipped with two roof-mounted, rotating 32 beam LiDAR sensors. Each LiDAR has a 40$^{\circ}$ vertical field of view, with 30$^{\circ}$ overlapping field of view and 50$^{\circ}$ total field of view with both LiDAR. LiDAR range is up to 200 meters, roughly twice the range as the sensors used in nuScenes and KITTI. On average, our LiDAR sensors produce a point cloud at each sweep with three times the density of the LiDAR sweeps in the nuScenes \cite{holger:nuscenes:2019} dataset (ours $\sim107,000$ points vs. nuScenes $\sim35,000$ points). The two LiDAR sensors rotate at 10 Hz and are out of phase, i.e. rotating in the same direction and speed but with an offset to avoid interference. Each 3D point is motion-compensated to account for ego-vehicle motion throughout the duration of the sweep capture. The vehicles have 7 high-resolution ring cameras ($1920 \times 1200$) recording at 30 Hz with overlapping fields of view, providing 360$^{\circ}$ coverage. In addition, there are 2 front-facing stereo cameras ($2056 \times 2464$ with a 0.2986 m baseline) sampled at 5 Hz. Faces and license plates are procedurally blurred in camera data to maintain privacy. Finally, 6-DOF localization for each timestamp comes from a combination of GPS-based and sensor-based localization. Vehicle localization and maps use a city-specific coordinate system described in more detail in the Appendix. Sensor measurements for particular driving sessions are stored in ``logs'', and we provide intrinsic and extrinsic calibration data for the LiDAR sensors and all 9 cameras for each log. Figure~\ref{fig:3D_vis} visualizes our sensor data in 3D. Similar to \cite{rangesh:LISAdataset:2017}, we place the origin of the ego-vehicle coordinate system at the center of the rear axle. All LiDAR data is provided in the ego-vehicle coordinate system, rather than in the respective LiDAR sensor coordinate frames. All sensors are roof-mounted, with a LiDAR sensor surrounded by 7 ``ring'' cameras (clockwise: facing front center, front right, side right, rear right, rear left, side left, and front left) and 2 stereo cameras. Figure~\ref{fig:sensors} visualizes the geometric arrangement of our sensors.

\subsection{Maps}
Argoverse contains three distinct map components -- (1) a vector map of lane centerlines and their attributes; (2) a rasterized map of ground height, and (3) a rasterized map of driveable area and region of interest (ROI).


\noindent \textbf{Vector Map of Lane Geometry}. Our \textit{vector map} consists of semantic road data represented as a localized graph rather than rasterized into discrete samples. The vector map we release is a simplification of the map used in fleet operations. In our vector map, we offer lane centerlines, split into lane segments. We observe that vehicle trajectories generally follow the center of a lane so this is a useful prior for tracking and forecasting.

A lane segment is a segment of road where cars drive in single-file fashion in a single direction. Multiple lane segments may occupy the same physical space (\eg in an intersection). Turning lanes which allow traffic to flow in either direction are represented by two different lanes that occupy the same physical space.

For each lane centerline, we provide a number of semantic attributes. These lane attributes describe whether a lane is located within an intersection or has an associated traffic control measure (Boolean values that are not mutually inclusive). Other semantic attributes include the lane's turn direction (\textit{left, right, or none}) and the unique identifiers for the lane's predecessors (lane segments that come before) and successors (lane segments that come after) of which there can be multiple (for merges and splits, respectively). Centerlines are provided as ``polylines'', \ie an ordered sequence of straight segments. Each straight segment is defined by 2 vertices: $(x_i,y_i,z_i)$ start and $(x_{i+1},y_{i+1},z_{i+1})$ end. Thus, curved lanes are approximated with a set of straight lines.

We observe that in Miami, lane segments that could be used for route planning are on average $3.84$ $\pm 0.89$ m wide. In Pittsburgh, the average width is $3.97$ $\pm 1.04$ m. Other types of lane segments that would not be suitable for self-driving, \eg bike lanes, can be as narrow as $0.97$ m in Miami and as narrow as $1.06$ m in Pittsburgh.


\noindent \textbf{Rasterized Driveable Area Map}.\label{driveable:area:description} Our maps include binary driveable area labels at 1 meter grid resolution. A driveable area is an area where it is possible for a vehicle to drive (though not necessarily legal). Driveable areas can encompass a road's shoulder in addition to the normal driveable area that is represented by a lane segment. We annotate 3D objects with track labels if they are within 5 meters of the driveable area (Section~\ref{label:Track_Annotations}). We call this larger area our \emph{region of interest} (ROI).

\begin{figure*}
\centering
\includegraphics[width=\linewidth]{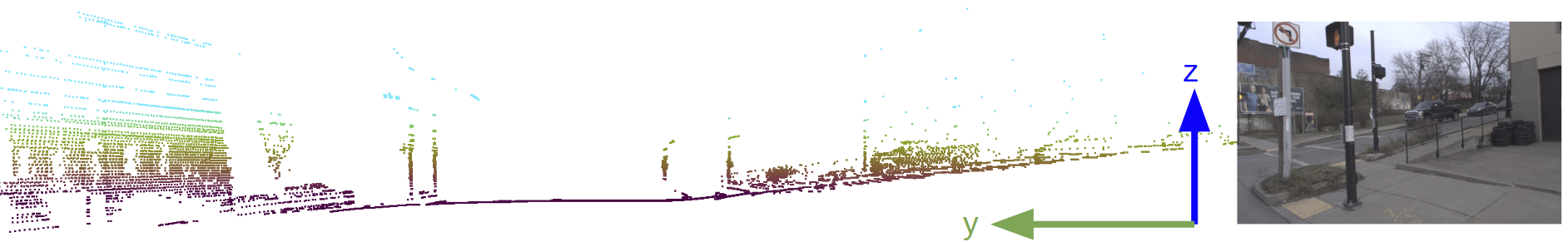}
\caption{\textbf{Uneven ground scene in Argoverse dataset.} Some Argoverse scenes contain uneven ground, which is challenging to remove with simple heuristics (e.g. assuming that ground is planar). Above, we show a LiDAR slice with a slope on the right side and corresponding front-right camera image.}\label{fig:ground_removal}
\end{figure*}

\noindent \textbf{Rasterized Ground Height Map}. Finally, our maps include real-valued ground height at 1 meter grid resolution. Knowledge of ground height can be used to remove LiDAR returns on static ground surfaces and thus makes the 3D detection of dynamic objects easier. Figure~\ref{fig:ground_removal} shows a cross section of a scene with uneven ground height.

\subsection{3D Track Annotations}
\label{label:Track_Annotations}



\begin{figure}
    \centering
        \includegraphics[width=1.0\linewidth]{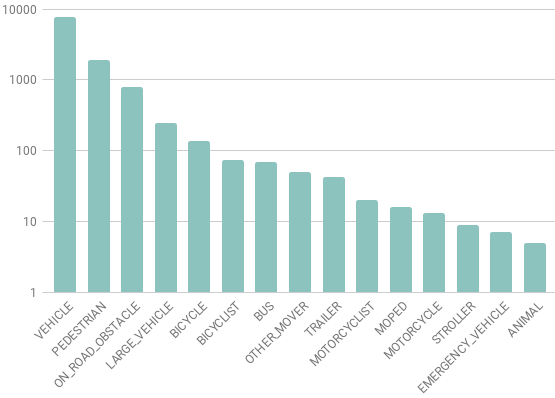}
    \caption{\textbf{Distribution of object classes.} This plot shows, in log scale, the number of 3D object tracks annotated for each class in the 113 log segments in the Argoverse 3D Tracking dataset.}
    \label{fig:classdistribution}
\end{figure}

The \emph{Argoverse Tracking Dataset} contains 113 vehicle log segments with human-annotated 3D tracks. These 113 segments vary in length from 15 to 30 seconds and collectively contain 11,052 tracked objects. We compared these with other datasets in Table~\ref{table:DB_compare}. For each log segment, we annotated all objects of interest (both dynamic and static) with bounding cuboids which follow the 3D LiDAR returns associated with each object over time. We only annotated objects within 5 m of the \emph{driveable area} as defined by our map. For objects that are not visible for the entire segment duration, tracks are instantiated as soon as the object becomes visible in the LiDAR point cloud and tracks are terminated when the object ceases to be visible. The same object ID is used for the same object, even if temporarily occluded.  
Each object is labeled with one of 15 categories, including ON\_ROAD\_OBSTACLE and OTHER\_MOVER for static and dynamic objects that do not fit into other predefined categories. More than 70\% of tracked objects are vehicles, but we also observe pedestrians, bicycles, mopeds, and more. Figure~\ref{fig:classdistribution} shows the distribution of classes for annotated objects. All track labels pass through a manual quality assurance review process. Figures~\ref{fig:teaser} and~\ref{fig:3D_vis} show qualitative examples of our human annotated labels. We divide our annotated tracking data into 65 training, 24 validation, and 24 testing sequences. 

\subsection{Mined Trajectories for Motion Forecasting} 
\label{label:forecasting_mining}
\begin{figure}
    \centering

    \includegraphics[width=\columnwidth]{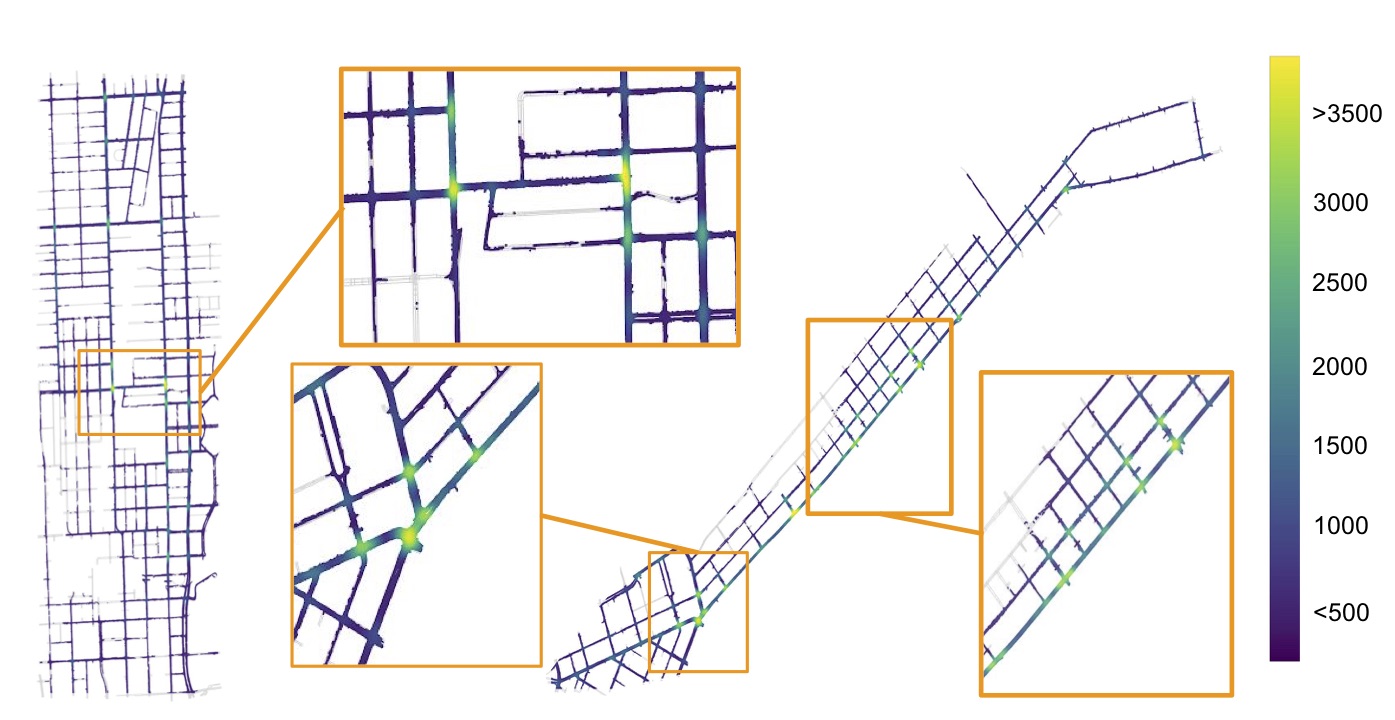}
    \caption{\textbf{Distribution of mined trajectories.} The colors indicate the number of mined trajectories across the maps of Miami (left) and Pittsburgh (right). The heuristics to find interesting vehicle behavior lead to higher concentrations in intersections and on busy roads such as Liberty and Penn Ave (southeast roads in bottom right inset).}
    \label{fig:map_tracked_vehicles}
\end{figure}

We are also interested in studying the task of \emph{motion forecasting} in which we predict the location of a tracked object some time in the future. Motion forecasts can be critical to safe autonomous vehicle motion planning. While our human-annotated 3D tracks are suitable training and test data for motion forecasting, the motion of many vehicles is relatively uninteresting -- in a given frame, most cars are either parked or traveling at nearly constant velocity. Such tracks are hardly a representation of real forecasting challenges. We would like a benchmark with more diverse scenarios e.g. managing an intersection, slowing for a merging vehicle, accelerating after a turn, stopping for a pedestrian on the road, etc. To sample enough of these \emph{interesting} scenarios, we track objects from 1006 driving hours across both Miami and Pittsburgh and find vehicles with interesting behavior in 320 of those hours. In particular, we mine for vehicles that are either (1) at intersections, (2) taking left or right turns, (3) changing to adjacent lanes, or (4) in dense traffic. In total, we collect 324,557 five second sequences and use them in the forecasting benchmark. Figure~\ref{fig:map_tracked_vehicles} shows the geographic distribution of these sequences. Each sequence contains the 2D, bird's eye view centroid of each tracked object sampled at 10 Hz. The ``focal'' object in each sequence is always a vehicle, but the other tracked objects can be vehicles, pedestrians, or bicycles. Their trajectories are available as context for ``social'' forecasting models. The 324,557 sequences are split into 205,942 train, 39,472 validation, and 78,143 test sequences. Each sequence has one challenging trajectory which is the focus of our forecasting benchmark. The train, validation, and test sequences are taken from disjoint parts of our cities, i.e. roughly one eighth and one quarter of each city is set aside as validation and test data, respectively. This dataset is far larger than what could be mined from publicly available autonomous driving datasets. While data of this scale is appealing because it allows us to see rare behaviors and train complex models, it is too large to exhaustively verify the accuracy of the mined trajectories and, thus, there is some noise and error inherent in the data.

\section{3D Object Tracking} \label{tracking_pipeline}


In this section, we investigate how various baseline tracking methods perform on the Argoverse 3D tracking benchmark. Our baseline methods utilize a hybrid approach with LiDAR and ring camera images and operate directly in 3D. In addition to measuring the baseline difficulty of our benchmark, we measure how simple map-based heuristics can influence tracking accuracy. For these baselines, we track and evaluate \emph{vehicles} only.

Given a sequence of $F$ frames, where each frame contains a set of ring camera images and 3D points from LiDAR $\{P_i \mid i = 1,...,N\}$, where $P_i$ $\in$ $\mathbb{R}^3$ of $x,y,z$ coordinates, we want to determine a set of track hypotheses $\{T_j \mid j=1,...,n\}$ where $n$ is the number of unique objects in the whole sequence, and $T_j$ contains the set of object center locations and orientation. We usually have a dynamic observer as our car is in motion more often than not. The tracked vehicles in the scene around us can be static or moving.

\textbf{Baseline Tracker}. Our baseline tracking pipeline clusters LiDAR returns in driveable region (labeled by the map) to detect potential objects, uses Mask R-CNN~\cite{maskrcnn} to prune non-vehicle LiDAR returns, associates clusters over time using nearest neighbor and the Hungarian algorithm, estimates transformations between clusters with iterative closest point (ICP), and estimates vehicle pose with a classical Kalman Filter using constant velocity motion model. The same predefined bounding box size is used for all vehicles. 

When no match can be found by Hungarian method for an object, the object pose is maintained using only motion model up to 5 frames before being removed or associated to a new cluster. This enables our tracker to maintain same object ID even if the object is occluded for a short period of time and reappears.  If a cluster is not associated with current tracked objects, we initialize a new object ID for it. 

The tracker uses the following map attributes:

\noindent \textbf{Driveable area.} Since our baseline is focused on vehicle tracking, we constrain our tracker to the driveable area as specified by the map. This driveable area covers any region where it is possible for the vehicle to drive (see Section \ref{driveable:area:description}). This constraint reduces the opportunities for false positives.

\noindent \textbf{Ground height.} We use map information to remove LiDAR returns on the ground. In contrast to local ground-plane estimation methods, the map-based approach is effective in sloping and uneven environments. 
    
\noindent \textbf{Lane Direction.}  Determining the vehicle orientation from LiDAR alone is a challenging task even for humans due to LiDAR sparsity and partial views. We observe that vehicle orientation rarely violates lane direction, especially so outside of intersections. Fortunately, such information is available in our dataset, so we adjust vehicle orientation based on lane direction whenever the vehicle is not at the intersection and contains too few LiDAR points. 
    
\subsection{Evaluation}

We leverage standard evaluation metrics commonly used for multiple object tracking (MOT) \cite{MOT16,Bernardin2008EvaluatingMO}. The MOT metric relies on a distance/similarity metric between ground truth and predicted objects to determine an optimal assignment. Once an assignment is made, we use three distance metrics for MOTP: MOTP-D (centroid distance), MOTP-O (orientation error), and MOTP-I (Intersection-over-Union error). MOTP-D is computed by the 3D bounding box centroid distance between associated tracker output and ground truth, which is also used in MOTA as detection association range. Our threshold for ``missed'' tracks is 2 meters, which is half of the average family car length in the US. MOTP-O is the smallest angle difference about the z (vertical) axis such that the front/back object
orientation is ignored, and MOTP-I is the amodal shape estimation error, computed by the $1-IoU$ of 3D bounding box after aligning orientation and centroid as in nuScenes~\cite{holger:nuscenes:2019}. For all three MOTP scores, lower scores indicate higher accuracy.

In our experiments, we run our tracker over the 24 logs in the \emph{Argoverse 3D Tracking} test set. 
We are also interested in the relationship between tracking performance and distance. We apply a threshold (30, 50, 100 m) to the distance between vehicles and our ego-vehicle and only evaluate annotations and tracker output within that range. The results in Table~\ref{Tab:tracking} show that our baseline tracker performs well at short range where the LiDAR sampling density is higher, but struggles for objects beyond 50 m.

\begin{figure}
     \subfloat[Without lane information] {
     \includegraphics[width=0.49\columnwidth]{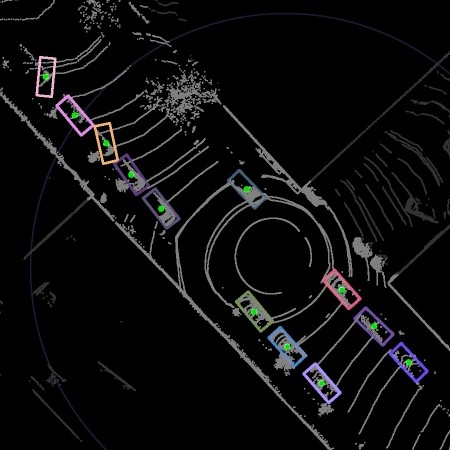}
     }
     \subfloat[With lane information] { 
     \includegraphics[width=0.49\columnwidth]{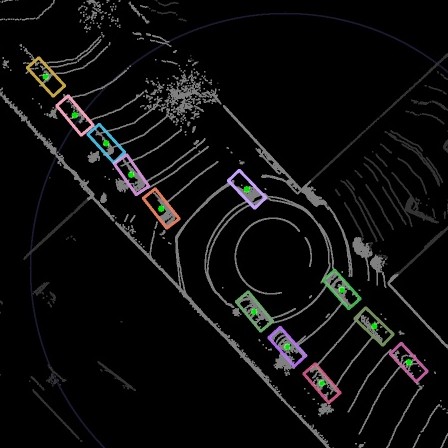}
     }
     \caption{\textbf{Tracking with orientation snapping.} Using lane direction information helps to determine the vehicle orientation for detection and tracking.}
     \label{fig:use_map_lane}
 \end{figure}

We compare our baseline tracker with three ablations that include: 1) using map-based ground removal and lane direction from the map; 2) using naive plane-fitting ground removal and lane direction from the map; 3) using map-based ground removal and no lane direction from the map. The results in Table~\ref{Tab:tracking2} show that map-based ground removal leads to better 3D IoU score and slightly better detection performance (higher MOTA) than a plane-fitting approach at longer ranges, but slightly worse orientation. On the other hand, lane direction from the map significantly improves orientation performance, as shown in Figure~\ref{fig:use_map_lane}.

We have employed relatively simple baselines to track objects in 3D. We believe that our data enables new approaches to map-based and multimodal tracking research. 




%
\begin{table*}[t]
\center
\begin{adjustbox}{max width=\textwidth}
\begingroup
\setlength{\tabcolsep}{10pt} 
\begin{tabular}{ c | c c c c c c c c c c c }
 \textsc{range} (m)  &  \textsc{MOTA} & \textsc{MOTP-D} & \textsc{MOTP-O}& \textsc{MOTP-I} & \textsc{IDF1} & \textsc{MT}(\%) & \textsc{ML}(\%) & \#\textsc{FP} & \#\textsc{FN} & \textsc{IDsw} & \#\textsc{frag} \\
    \midrule 

 30   & \bf{65.5} & \bf{0.71} & 15.3 & \bf{0.25} & \bf{0.71} & \bf{0.67} & \bf{0.18} & \bf{5739} & \bf{10098} & \bf{356} & \bf{380}\\
 
 50  & 50.0 & 0.81 & 13.5 & 0.26 & 0.59 & 0.30  & 0.31 & 8191 & 30468 & 607 & 691\\
100   &  34.2 & 0.82 & \bf{13.3} & \bf{0.25} & 0.46 & 0.13 & 0.51 & 9225 & 66234 & 679 & 773 \\

     \bottomrule 
    \end{tabular}
\endgroup
\end{adjustbox}
\caption{\textbf{Tracking accuracy at different ranges using map for ground removal and orientation initialization.} From top to bottom, accuracy for vehicles within 30 m, 50 m, and 100 m.}
\label{Tab:tracking}
\end{table*}

\begin{table*}[t]
\center
\begin{adjustbox}{max width=\textwidth}
\begingroup
\setlength{\tabcolsep}{10pt} 
\begin{tabular}{ c | c |c|c c c c }
\textsc{range}  & \textsc{use} &  \textsc{ground} & \textsc{MOTA} & \textsc{MOTP-D} & \textsc{MOTP-O}& \textsc{MOTP-I} \\
(m) & \textsc{map lane} & \textsc{removal} & & (m) & (degree) & \\

     \midrule 
    & Y & map          & 65.5 & \bf{0.71} & 15.3  & \bf{0.25} \\
 30 & Y & plane-fitting&  \bf{65.8} & 0.72 & \bf{13.7} & 0.29 \\
    & N & map          & 65.4 &  \bf{0.71} & 25.3 & \bf{0.25} \\


     \midrule 
    & Y & map          &  \bf{50.0} & \bf{0.81} & 13.5  & \bf{0.26} \\
 50 & Y & plane-fitting& 49.3 & \bf{0.81} & \bf{12.5} & 0.29 \\ 
    & N & map          & 49.8 & \bf{0.81} & 27.7 & \bf{0.26} \\
    
    \midrule 
    &  Y & map           & \bf{34.2} & \bf{0.82} & 13.3 & \bf{0.25} \\
100 &  Y & plane-fitting & 33.6 & \bf{0.82} & \bf{12.5} & 0.28 \\
    &  N & map           & 34.1 & \bf{0.82} & 27.7 & \bf{0.25} \\

     \bottomrule 
    \end{tabular}
\endgroup
\end{adjustbox}
\caption{\textbf{Tracking accuracy comparison at different ranges while using different map attributes.} From top to bottom, accuracy for vehicles within 30 m, 50 m, and 100 m.}
\label{Tab:tracking2}
\end{table*}

\section{Motion Forecasting} \label{forecasting_pipeline}
In this section, we describe our pipeline for motion forecasting baselines.  

\textbf{1. Preprocessing}: As described in Section \ref{label:forecasting_mining}, we first mine for ``interesting'' sequences where a ``focal'' vehicle is observed for 5 seconds. As context, we have the centroids of all other tracked objects (including the AV itself) which are collapsed into one ``other'' class.

\textbf{Forecasting Coordinate System and Normalization}. The coordinate system we used for trajectory forecasting is a top-down, bird's eye view (BEV). There are three reference coordinate frames of interest to forecasting: (1) The raw trajectory data is stored and evaluated in the \emph{city} coordinate system (See Section \ref{supp:traj-mining-details} of the Appendix). (2) For models using lane centerlines as a reference path, we defined a \emph{2D curvilinear coordinate system} with axes tangential and perpendicular to the lane centerline. (3) For models without the reference path (without a map), we normalize trajectories such that the observed portion of the trajectory starts at the origin and ends somewhere on the positive x axis.
If $(x_i^t, y_i^t)$ represent coordinates of trajectory $V_i$ at timestep $t$, then this normalization makes sure that $y_i^{T_{obs}}=0$, where $T_{obs}$ is last observed timestep of the trajectory (Section \ref{section:problem_description}). We find this normalization works better than leaving trajectories in absolute map coordinates or absolute orientations. 

\textbf{2. Feature Engineering}: We define additional features to capture social or spatial context. For social context, we use the minimum distance to the objects in front, in back, and the number of neighbors. Such heuristics are meant to capture the social interaction between vehicles.  For spatial context, we use the map as a prior by computing features in the lane segment coordinate system. We compute the lane centerline corresponding to each trajectory and then map $(x_i^t,y_i^t)$ coordinates to the distance along the centerline $(a_i^t)$ and offset from the centerline $(o_i^t)$. In the subsequent sections, we denote social features and map features for trajectory $V_i$ at timestep $t$ by $s_i^t$ and $m_i^t$, respectively.

\textbf{3. Prediction Algorithm}: We implement Constant Velocity, Nearest Neighbor, and LSTM Encoder-Decoder based \cite{park:2018:rnntrajpred, graves:2013:seqgenrnn, sutskever:nips:2014} models using different combinations of features. The results are analyzed in Section \ref{section:forecasting_results}.


\subsection{Problem Description} \label{section:problem_description}
The forecasting task is framed as: \textit{given the past input coordinates of a vehicle trajectory} $V_i=(X_i,Y_i)$ \textit{where} $X_i = (x_i^t, y_i^t)$ \textit{for time steps} $t = \{1, \hdots, T_{obs}\}$, \textit{predict the future coordinates} $Y_i = (x_i^t,y_i^t)$ \textit{for time steps} $\{t = T_{obs + 1}, \hdots, T_{pred}\}$. For a car, 5 s is sufficient to capture the salient part of a trajectory, \eg. crossing an intersection. 
In this paper, we define the motion forecasting task as observing 20 past frames (2 s) and then predicting 30 frames (3 s) into the future. Each forecasting task can leverage the trajectories of other objects in the same sequence to capture the social context and map information for spatial context.
 
\subsection{Evaluation of Multiple Forecasts} \label{section:forecasting_multimodal}
Predicting the future is difficult. Often, there are several plausible future actions for a given observation. In the case of autonomous vehicles, it is important to predict \emph{many} plausible outcomes and not simply the \emph{most likely} outcome. While some prior works have evaluated forecasting in a deterministic, unimodal way, we believe a better approach is to follow the evaluation methods similar to DESIRE \cite{desire_LeeCVCTC17}, Social GAN \cite{Gupta_2018_CVPR}, R2P2 \cite{rhinehart2018r2p2} and \cite{rhinehart2019precog} wherein they encourage algorithms to output multiple predictions. Among the variety of metrics evaluated in \cite{rhinehart2018r2p2} was the \textit{minMSD} over $K$ \textit{number of samples} metric, where $K=12$. A similar metric is used in \cite{desire_LeeCVCTC17} where they allow $K$ to be up to 50.  
We follow the same approach and use minimum Average Displacement Error (minADE) and minimum Final Displacement Error (minFDE) over $K$ predictions as our metrics, where $K=1,3,6,9$. Note that minADE refers to ADE of the trajectory which has minimum FDE, and not minimum ADE, since we want to evaluate the single best forecast. That said, minADE error might not be a sufficient metric. As noted in \cite{rhinehart2018r2p2} and \cite{rhinehart2019precog}, metrics like minMSD or minFDE can only evaluate how good is the best trajectory, but not how good are all the trajectories. A model having 5 good trajectories will have the same error as the model having 1 good and 4 bad trajectories. Further, given the multimodal nature of the problem, it might not be fair to evaluate against a single ground truth. In an attempt to evaluate based on the quality of predictions, we propose another metric: Drivable Area Compliance (DAC). If a model produces $n$ possible future trajectories and $m$ of those exit the drivable area at some point, the DAC for that model would be $(n-m)/n$. Hence, higher DAC means better quality of forecasted trajectories. Finally, we also use Miss Rate (MR) \cite{yeh2019diverse} with a threshold of 1.0 meter. It is again a metric derived from the distribution of final displacement errors. If there are $n$ samples and $m$ of them had the last coordinate of their best trajectory more than 2.0 m away from ground truth, then miss rate is $m/n$.
The map-based baselines that we report have access to a semantic vector map. As such, they can generate \textit{K} different hypotheses based on the branching of the road network along a particular observed trajectory. We use centerlines as a form of hypothetical reference paths for the future. Our  heuristics generate $K=10$ centerlines. Our map gives us an easy way to produce a compact yet diverse set of forecasts. 
Nearest Neighbor baselines can further predict variable number of outputs by considering different number of neighbors.


\subsection{Results} \label{section:forecasting_results}

In this section, we evaluate the effect of multimodal predictions, social context, and spatial context (from the vector map) to improve motion forecasting over  horizons of 3 seconds into the future. We evaluated the following models:
\begin{itemize}[noitemsep,nolistsep]
    \item \textit{Constant Velocity}: Compute the mean velocity $(v_{xi}, v_{yi})$ from $t=\{1,\hdots, T_{obs}\}$ and then forecast $(x_i^t, y_i^t)$ for $t=\{T_{obs+1},\hdots, T_{pred}\}$  using $(v_{xi}, v_{yi})$ as the constant velocity.
    \item \textit{NN}: Nearest Neighbor regression where trajectories are queried by $(x_i^t, y_i^t)$ for $t=\{1,\hdots, T_{obs}\}$. To make $K$ predictions, we performed a lookup for $K$ Nearest Neighbors.
    \item \textit{LSTM}: LSTM Encoder-Decoder model where the input is $(x_i^t, y_i^t)$ for $t=\{1,\hdots, T_{obs}\}$ and output is $(x_i^t, y_i^t)$ for $t=\{T_{obs+1},\hdots, T_{pred}\}$. This is limited to one prediction because we used a deterministic model.
    \item \textit{LSTM+social}: Similar to \textit{LSTM} but with input as $(x_i^t, y_i^t, s_i^t)$, where $s_i^t$ denotes social features.
    \item \textit{NN+map(prune)}: This baseline builds on $NN$ and prunes the number of predicted trajectories based on how often they exit the drivable area. Accordingly,  this method prefers predictions which are qualitatively good, and not just Nearest Neighbors.  
    \item \textit{NN+map(prior) m-G,n-C}: Nearest Neighbor regression where trajectories are queried by $(a_i^t, o_i^t)$ for $t=\{1,\hdots, T_{obs}\}$. m-G, n-C refers to $m$ guesses (m-G) allowed along each of $n$ different centerlines (n-C). Here, $m>1$, except when $K=1$.
    \item \textit{NN+map(prior) 1-G,n-C}: This is similar to the previous baseline. The only difference is that the model can make only 1 prediction along each centerline.
    \item \textit{LSTM+map(prior) 1-G,n-C}: Similar to \textit{LSTM} but with input as $(a_i^t, o_i^t, m_i^t)$ and output as $(a_i^t, o_i^t)$, where $m_i^t$ denotes the map features obtained from the centerlines. Distances $(a_i^t, o_i^t)$ are then mapped to $(x_i^t, y_i^t)$ for evaluation. Further, we make only one prediction along each centerline because we used a deterministic model.
\end{itemize}

\begin{table*}[t]
\center
\begin{adjustbox}{max width=\textwidth}
\begingroup
\setlength{\tabcolsep}{10pt} 
\begin{tabular}{lcccc|cccc|cccc} 
&\multicolumn{4}{c}{\textsc{K=1} }
&  \multicolumn{4}{c}{\textsc{K=3} }
&  \multicolumn{4}{c}{\textsc{K=6} }
\\
\textsc{Baseline} & minADE $\downarrow$ & minFDE $\downarrow$ & DAC $\uparrow$ & MR $\downarrow$ & minADE $\downarrow$ & minFDE $\downarrow$ & DAC $\uparrow$ & MR $\downarrow$ & minADE $\downarrow$ & minFDE $\downarrow$ & DAC $\uparrow$ & MR $\downarrow$ \\ 
 \midrule
Constant Velocity  & 3.53 & 7.89 & 0.88 & 0.83 & -& -& -& -& -& -& -& -   \\
NN  & 3.45 & 7.88 & 0.87 & 0.87 & 2.16 & 4.53 & 0.87 & 0.70 & 1.71 & 3.29 & 0.87 & 0.54 \\
LSTM & \textbf{2.15} & 4.97&	0.93 & \textbf{0.75} & -& -& -& -& -& -& -& -  \\
LSTM+social & \textbf{2.15} &	\textbf{4.95}&	0.93 & \textbf{0.75} & -& -& -& -& -& -& -& -  \\
NN+map(prune) & 3.38 &	7.62 &	\textbf{0.99} & 0.86 &	\textbf{2.11}&	\textbf{4.36}&	\textbf{0.97}&	0.68&	\textbf{1.68}&	\textbf{3.19}&	0.94&	\textbf{0.52} \\
NN+map(prior) m-G,n-C &3.65&	8.12&	0.83&	0.94&	2.46&	5.06&	\textbf{0.97}&	0.63&	2.08&	4.02&	\textbf{0.96}&	0.58 \\
NN+map(prior) 1-G,n-C & 3.65&	8.12&	0.83&	0.94&	3.01&	6.43&	0.95&	0.80&	2.6&	5.32&	0.92&	0.75\\
LSTM+map(prior) 1-G,n-C & 2.92&	6.45&	0.98&	\textbf{0.75}&	2.31&	4.85&	0.97&	0.71&	2.08&	4.19&	0.95&	0.67\\
\bottomrule
\end{tabular}
\endgroup
\end{adjustbox}
\caption{Motion Forecasting Errors for different number of predictions. minADE: Minimum Average Displacement Error, minFDE: Minimum Final Displacement Error, DAC: Drivable Area Compliance, MR: Miss Rate (with a threshold of 2 m). Please refer to Section \ref{section:forecasting_multimodal} for definitions of these metrics ($\downarrow$ indicates lower is better).
}
\label{table:forecasting_results}
\end{table*} 

\begin{center}

\begin{figure}
    \centering

    \includegraphics[width=0.75\linewidth]{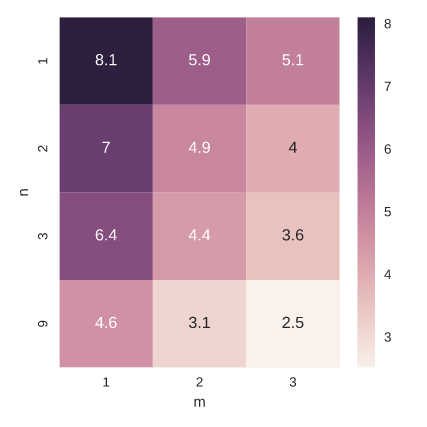}
    \caption{minFDE for \textit{NN+map(prior) m-G,n-C} with different values of \textit{n} (\#Centerlines) and \textit{m} (\#Predictions along each centerline). There's a trade-off between number of reference paths (\textit{n}) and number of predictions along each reference path (\textit{m}). Increasing \textit{n} ensures that we are capturing different high level scenarios while increasing \textit{m} makes sure we are capturing different velocity profiles along a given reference path. If the number of centerlines are enough, then for the same total number of predictions it is often better to make multiple predictions along fewer centerlines than to make 1 prediction along more centerlines.}
    \label{fig:forecasting_cl_heatmap}
\end{figure}

\end{center}

The results of these baselines are reported in Table \ref{table:forecasting_results}. 
When only 1 prediction is allowed, NN based baselines suffer from inaccurate neighbors and have poor minADE and minFDE. On the other hand, LSTM based baselines are able to at least learn the trajectory behaviors and have better results. $LSTM$ baselines with no map are able to obtain the best minADE and mindFDE for $K=1$. Also, baselines which use map as prior have a much higher DAC.
Now, as $K$ increases, $NN$ benefits from the map prior and consistently produces better predictions. When map is used for pruning, it further improves the selected trajectories and provides the best minADE and minFDE.
\textit{LSTM+map(prior) 1-G,n-C} outperforms \textit{NN+map(prior) 1-G,n-C} highlighting the fact that LSTM does a better job generalizing to curvilinear coordinates. Further, using the map as a prior always provides better DAC, proving that our map helps in forecasting trajectories that follow basic map rules like staying in the driveable area.
Another interesting comparison is between \textit{NN+map(prior) 1-G,n-C} and \textit{NN+map(prior) m-G,n-C}. The former comes up with many reference paths (centerlines) and makes one prediction along each of those paths. The latter comes up with fewer reference paths but produces multiple predictions along each of those paths. The latter outperforms the former in all 3 metrics, showing the importance of predicting trajectories which follow different velocity profiles along the same reference paths. Figure \ref{fig:forecasting_cl_heatmap} reports the results of an ablation study for different values of \textit{m} and \textit{n}.
Finally, when having access to HD vector maps and being able to make multiple predictions ($K=6)$, even a shallow model like \textit{NN+map(prior) m-G,n-C} is able to outperform a deterministic deep model \textit{LSTM+social} ($K=1$) which has access to social context.

\begin{figure}
\centering

\includegraphics[width=0.49\linewidth]{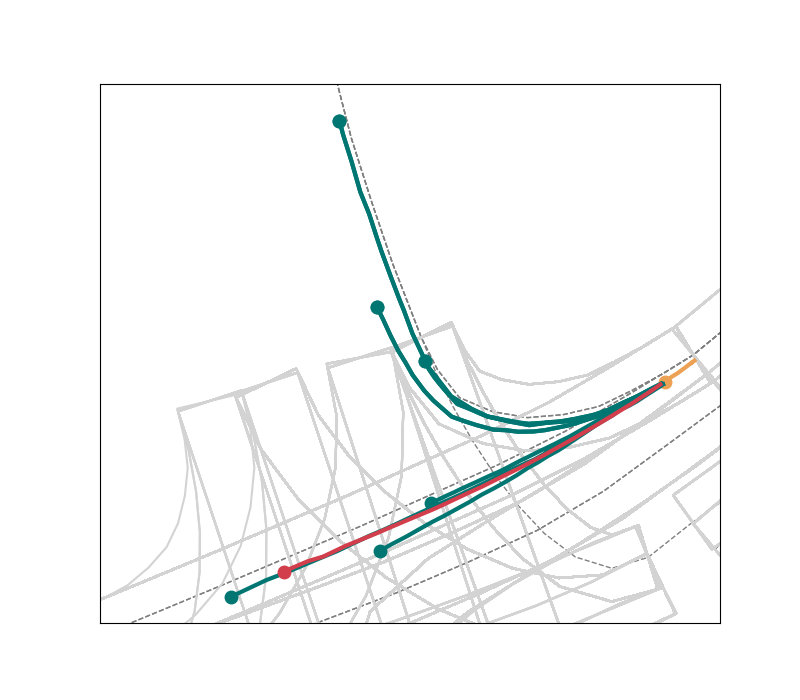}
\includegraphics[width=0.49\linewidth]{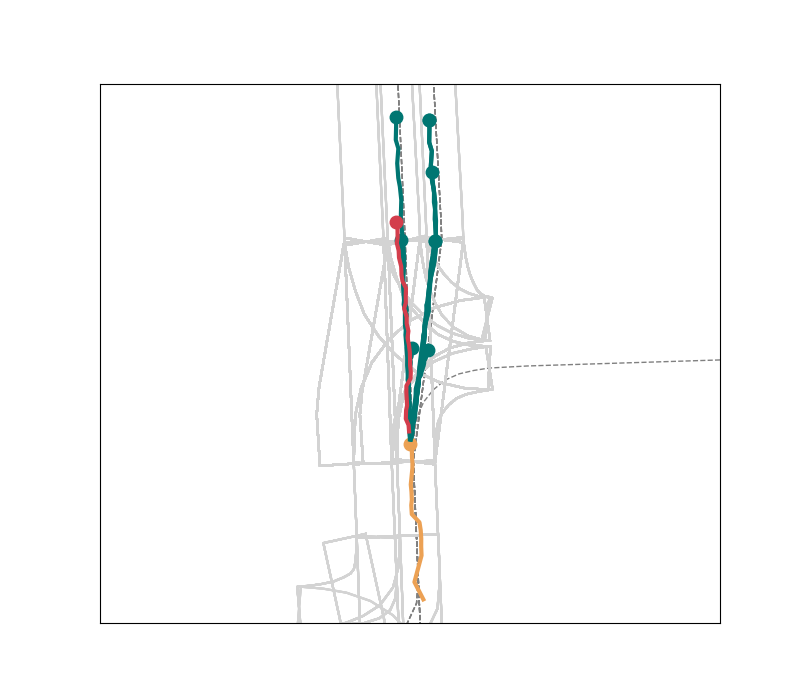}\\

\includegraphics[width=0.49\linewidth,trim={0 1mm 0 0},clip]{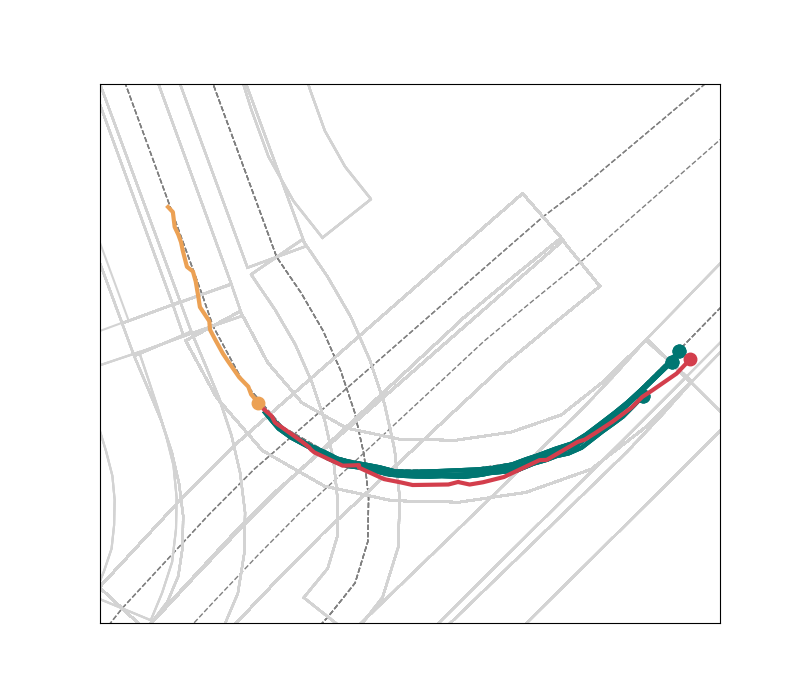}
\includegraphics[width=0.49\linewidth]{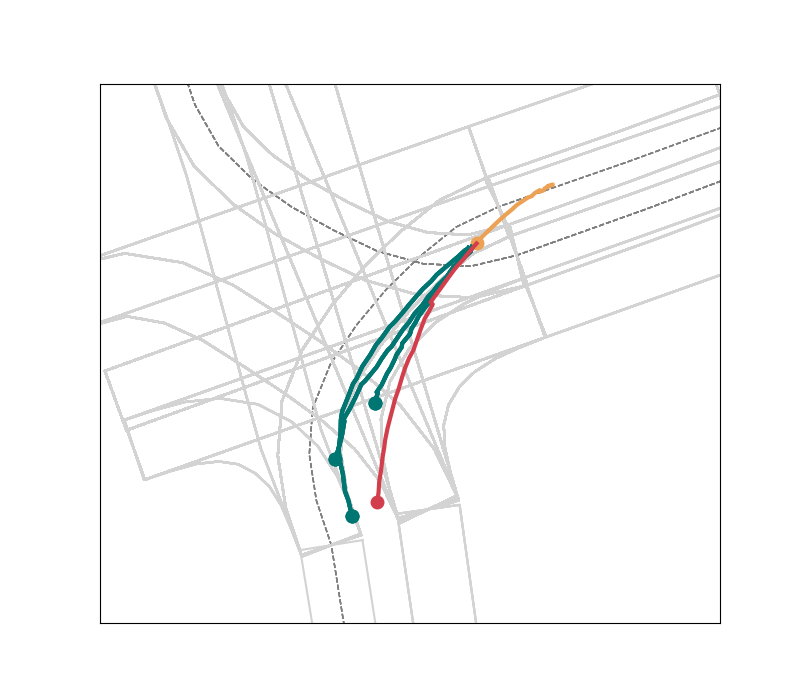}

\caption{\textbf{Qualitative results from \textit{NN+map(prior) m-G,n-C} motion forecasting baseline.} The orange trajectory represents the observed 2 s. Red represents ground truth for the next 3 seconds and green represents the multiple forecasted trajectories for those 3 s. \textbf{Top left}: The car starts to accelerate from a stop line and the model is able to predict 2 different modes (right turn and go straight) along with different velocity profiles along those modes. \textbf{Top right}: The model is able to predict 2 different scenarios -- lane change and staying in the same lane. \textbf{Bottom left}: The model is able to cross a complex intersection and take a wide left turn without violating any lane rules because it is able to use the vector map to generate a reference path. \textbf{Bottom right}: The predictions account for different ways in which a left turn can be taken, in terms of velocity profile and turning radius.}\label{fig:forecasting_results}
\end{figure}
\section{Discussion} \label{section:discussion}
Argoverse represents two large-scale datasets for autonomous driving research. The Argoverse datasets are the first such datasets with rich map information such as lane centerlines, ground height, and driveable area. We examine baseline methods for 3D tracking with map-derived context. We also mine one thousand hours of fleet logs to find diverse, real-world object trajectories which constitute our motion forecasting benchmark. We examine baseline forecasting methods and verify that map data can improve accuracy. We maintain a public leaderboard for 3D object tracking and motion forecasting. The sensor data, map data, annotations, and code which make up Argoverse are available at \mbox{our website \emph{Argoverse.org}}.

\textbf{Acknowledgements}. We thank our Argo AI colleagues for their invaluable assistance in supporting Argoverse. 

\small{Patsorn Sangkloy is supported by a a Royal Thai Government Scholarship. James Hays receives research funding from Argo AI, which is developing products related to the research described in this paper. In addition, the author serves as a Staff Scientist to Argo AI. The terms of this arrangement have been reviewed and approved by Georgia Tech in accordance with its conflict of interest policies.}


{\small
\bibliographystyle{ieee_fullname}
\bibliography{argoversearxiv}
}

\appendix

\section*{Appendices}

We present additional details about our map (Appendix \ref{supp:map-details}), our 3D tracking taxonomy of classes (Appendix \ref{supp:class-details}), our trajectory mining (Appendix \ref{supp:traj-mining-details}), and our 3D tracking algorithm (Appendix \ref{supp:trackingdetails}). 


\section{Supplemental Map Details}
\label{supp:map-details}

In this appendix, we describe details of our map coordinate system and the functions exposed by our map API, and we visualize several semantic attributes of our vector map. 
Our map covers 204 linear kilometers of lane centerlines in Miami and 86 linear kilometers in Pittsburgh. In terms of driveable area, our map covers 788,510 $\mbox{m}^2$ in Miami and 286,104 $\mbox{m}^2$ in Pittsburgh.

\subsection{Coordinate System}
The model of the world that we subscribe to within our map and dataset is a local tangent plane centered at a central point located within each city. This model has a flat earth assumption which is approximately correct at the scale of a city. Thus, we provide map object pose values in city coordinates. City coordinates can be converted to the UTM (Universal Transverse Mercator) coordinate system by simply adding the city's origin in UTM coordinates to the object's city coordinate pose. The UTM model divides the earth into 60 flattened, narrow zones, each of width 6 degrees of longitude. Each zone is segmented into 20 latitude bands. In Pittsburgh, our city origin lies at 583710.0070 Easting, 4477259.9999 Northing in UTM Zone 17. In Miami, our city origin lies at 
580560.0088 Easting, 2850959.9999 Northing in UTM Zone 17.

We favor a city-level coordinate system because of its high degree of interpretability when compared with geocentric reference coordinate systems such as the 1984 World Geodetic System (WGS84). While WGS84 is widely used by the Global Positioning System, the model is difficult to interpret at a city-scale; because its coordinate origin is located at the Earth's center of mass, travel across an entire city corresponds only to pose value changes in the hundredth decimal place.  The conversion back and forth between UTM and WGS84 is well-known and is documented in detail in \cite{Snyder:1987}.

We provide ground-truth object pose data in the ego-vehicle frame, meaning a single SE(3) transform is required to bring points into the city frame for alignment with the map:
$$
p_{city} = ({}^{city} T_{egovehicle}) \mbox{ } (p_{egovehicle})
$$

\begin{figure*}[h]

    \subfloat[]{
    \includegraphics[width=0.28\textwidth]{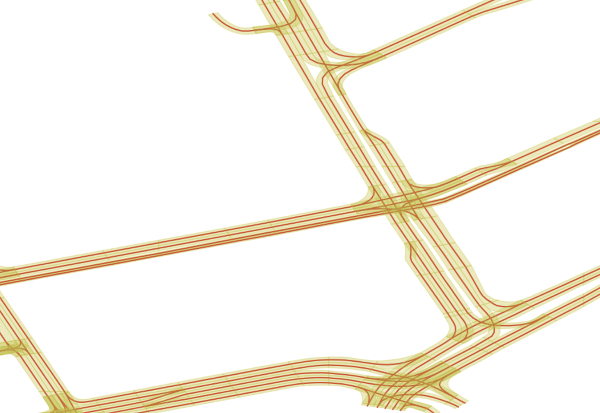}
    }
    \subfloat[]{
    \includegraphics[width=0.28\textwidth]{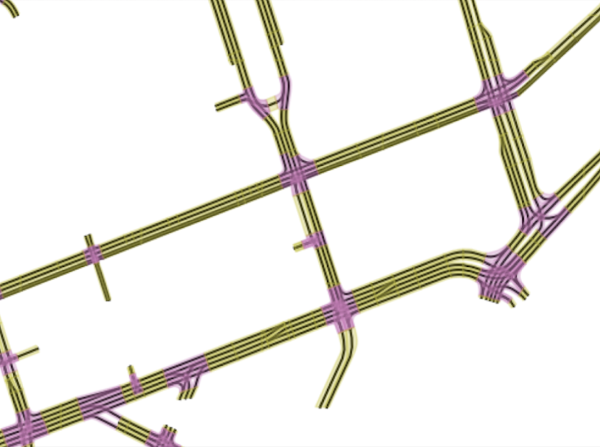}
    }
    \subfloat[]{
    \includegraphics[width=0.2\textwidth]{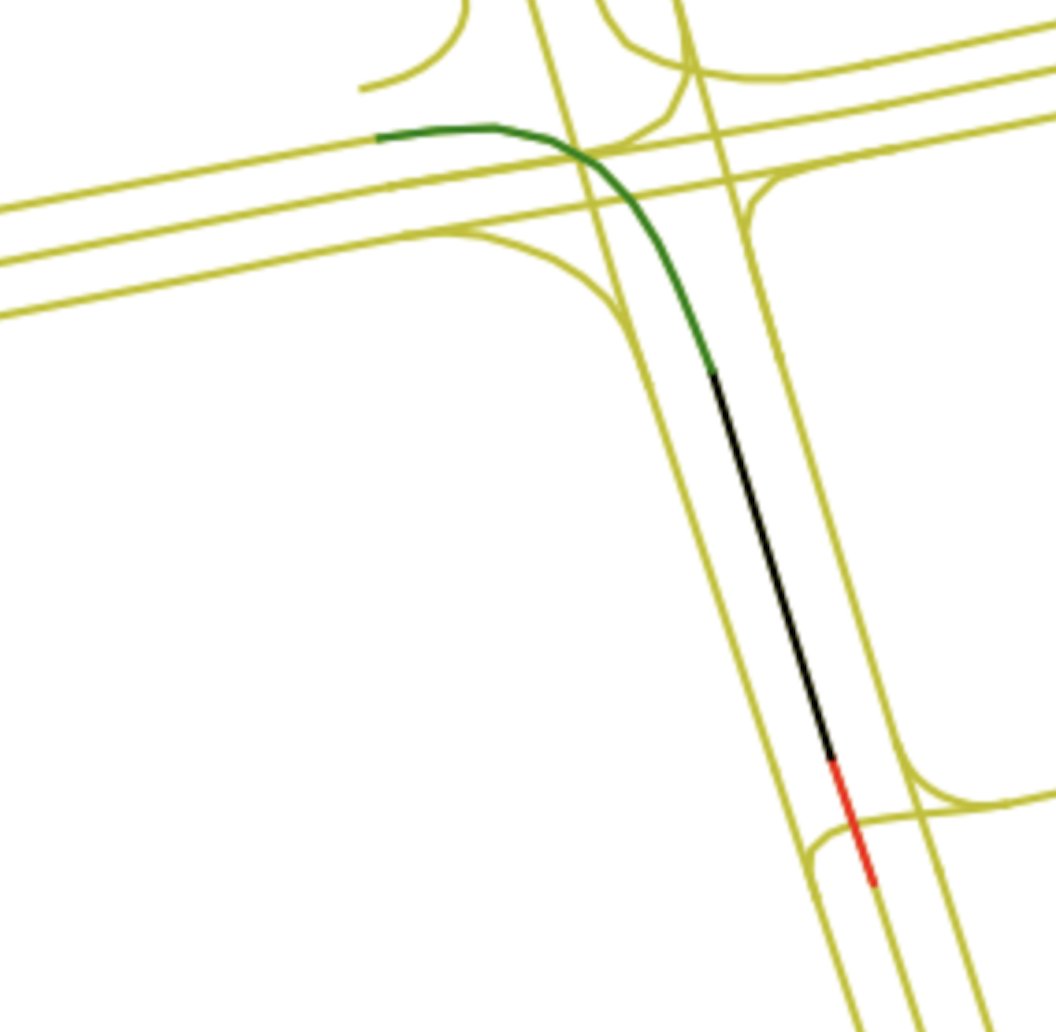}
    \includegraphics[width=0.2\textwidth]{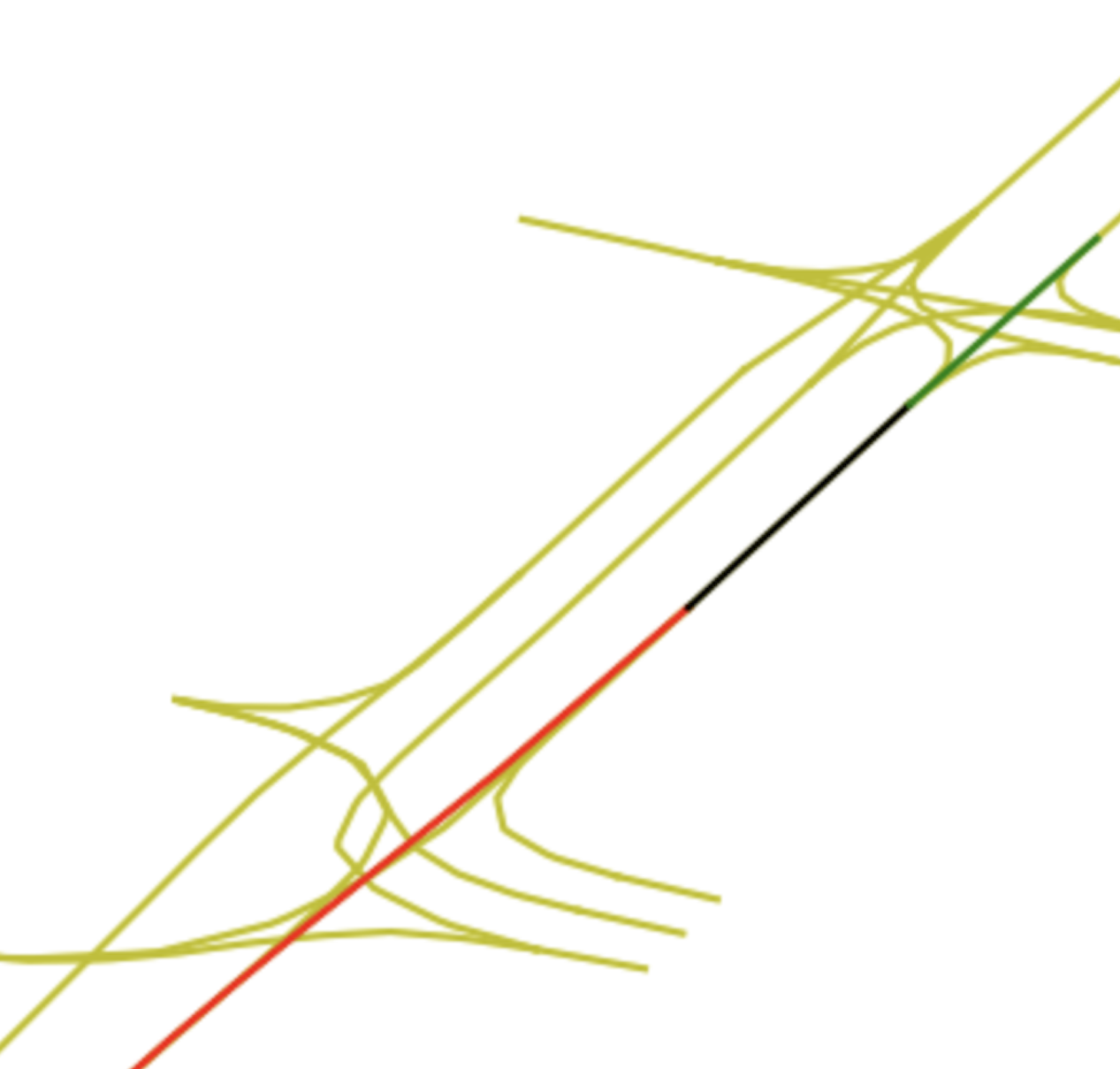}
    }
    \caption {(a) Lane centerlines and hallucinated area are shown in red and yellow, respectively. We provide \emph{lane} centerlines in our dataset because simple \emph{road} centerline representations cannot handle the highly complicated nature of real world mapping, as shown above with divided roads. (b) We show lane segments within intersections in pink, and all other lane segments in yellow. Black shows lane centerlines. (c) Example of a specific lane centerline's  successors and predecessors. Red shows the predecessor, green shows the successor, and black indicates the centerline segment of interest. }\label{fig:centerlines}
\end{figure*}

\begin{figure}
\centering
\subfloat[Lane geometry and connectivity]{
  \includegraphics[width=0.8\linewidth]{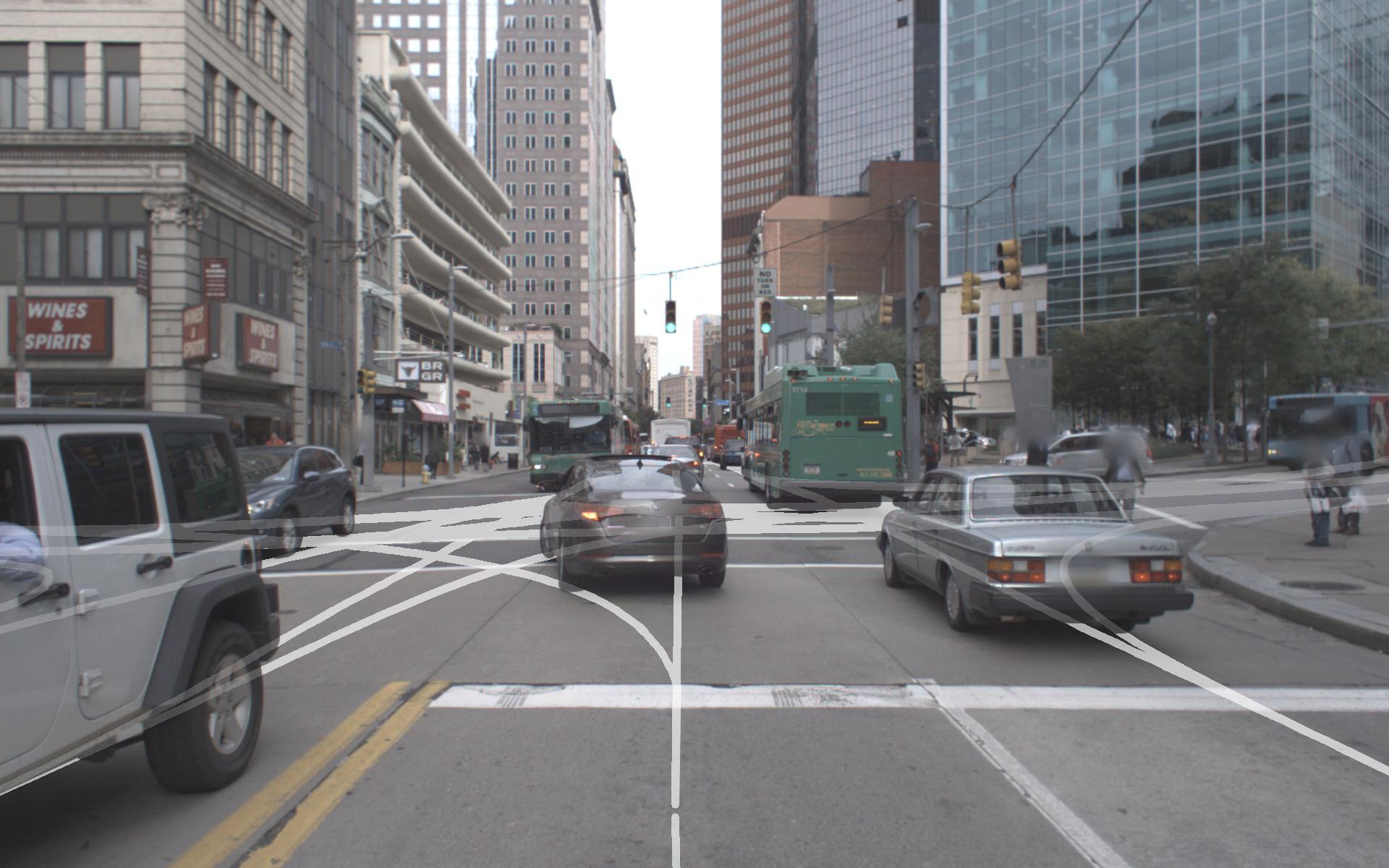}
}
\hspace{0mm}
\subfloat[Driveable area]{
   \includegraphics[width=0.8\linewidth]{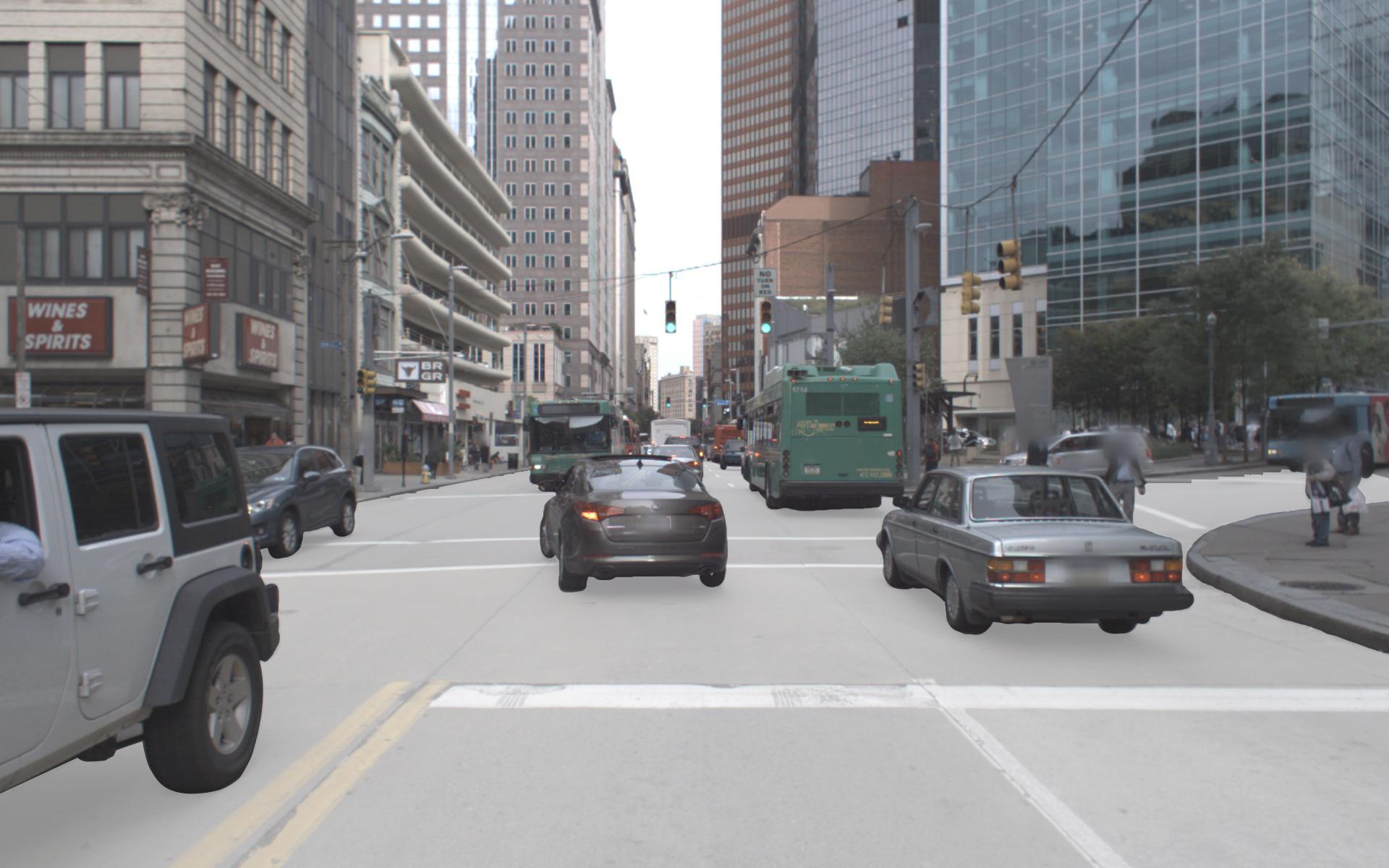}
}
\hspace{0mm}
\subfloat[Ground height]{
  \includegraphics[width=0.8\linewidth]{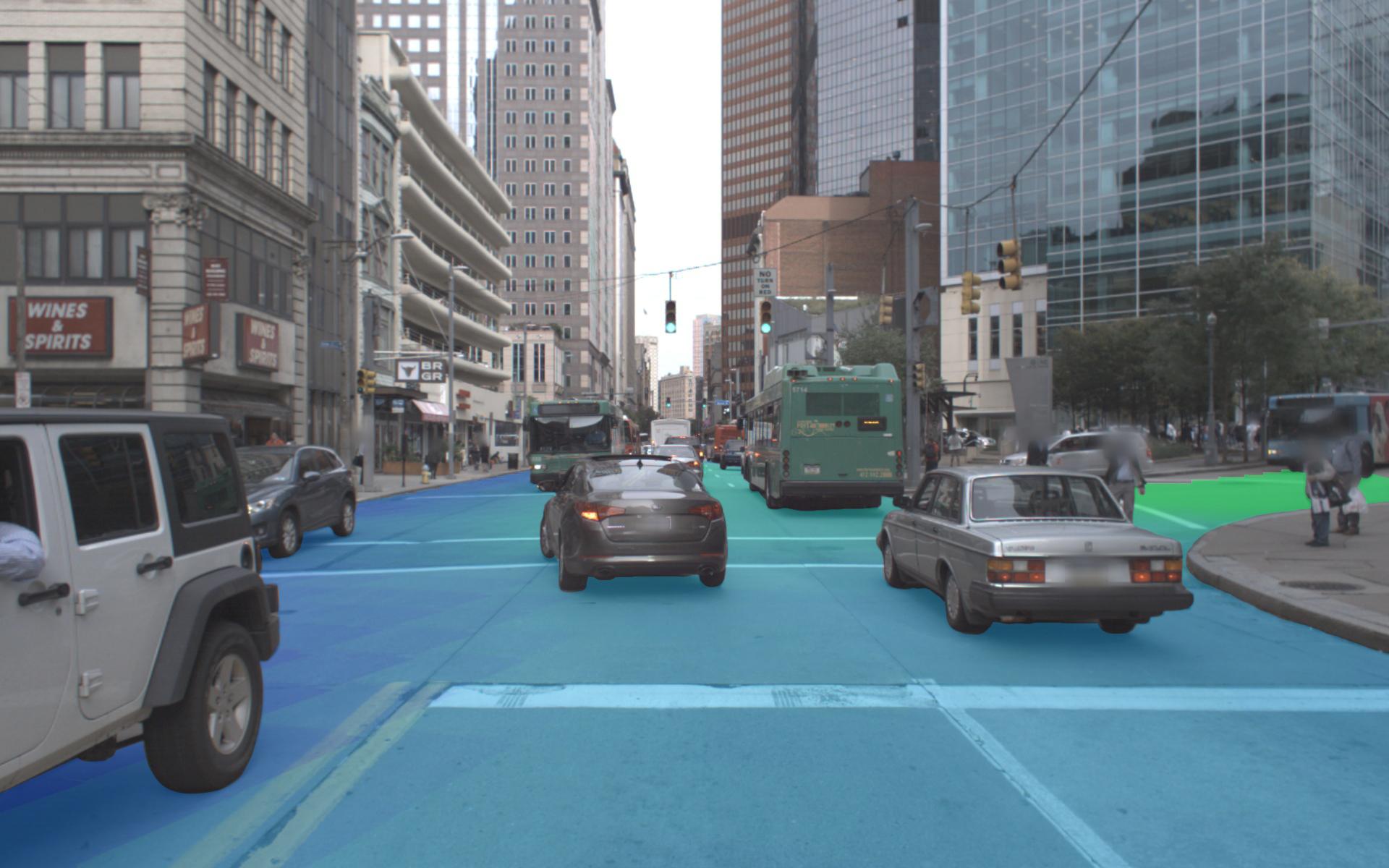}
}
 \caption{Examples of centerlines, driveable area, and ground height projected onto a camera image.}\label{fig:map_to_image_projection}
\end{figure}

Figure~\ref{fig:centerlines} shows examples of the centerlines which are the basis of our vector map. Centerline attributes include whether or not lane segments are in an intersection, and which lane segments constitute their predecessors and successors. Figure ~\ref{fig:map_to_image_projection} shows examples of centerlines, driveable area, and ground height projected onto a camera image.

\begin{figure*}
    \centering
    \includegraphics[width=\linewidth]{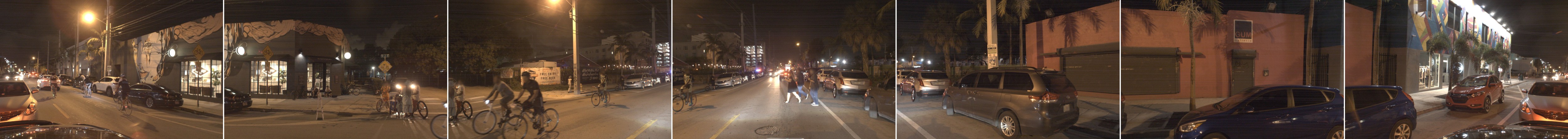}
    \includegraphics[width=\linewidth]{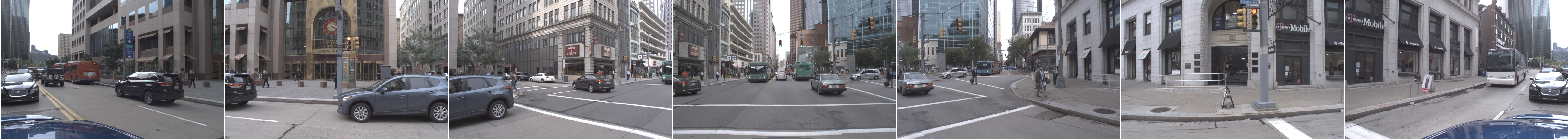}

    \caption{\textbf{Ring Camera Examples.} Scenes captured in Miami, Florida, USA (top) and Pittsburgh, Pennsylvania, USA (bottom) with our ring camera. Each row consists of 7 camera views with overlapping fields of view. Camera order is \textit{rear\_left, side\_left, front\_left, front\_center, front\_right, side\_right, rear\_right}.} 
       \label{fig:ring_camera:panorama:miami}
\end{figure*}

\subsection{Map API and Software Development Kit}



\begin{figure*}[b]
\begin{center}
 \begin{tabular}{||l l ||} 
 \hline
 \textbf{Function name} & \textbf{Description} \\ [0.5ex] 
 \hline\hline
 \texttt{remove\_non\_driveable\_area\_points} & 
Uses rasterized driveable area ROI to decimate LiDAR point cloud to \\ & only ROI points.
 \\
  \hline
 \texttt{remove\_ground\_surface} & 
Removes all 3D points within 30 cm of the ground surface.
 \\
 \hline
 \texttt{get\_ground\_height\_at\_xy} & 
Gets ground height at provided (x,y) coordinates.
 \\
 \hline
 \texttt{render\_local\_map\_bev\_cv2} & Renders a  Bird's Eye View (BEV) in OpenCV. \\
  \hline
 \texttt{render\_local\_map\_bev\_mpl} & Renders a Bird's Eye View (BEV) in Matplotlib. \\
  \hline
 \texttt{get\_nearest\_centerline} & Retrieves nearest lane centerline polyline. \\
  \hline
 \texttt{get\_lane\_direction} & Retrieves most probable tangent vector $\in \mathbbm{R}^2$ to lane centerline.\\
  \hline
 \texttt{get\_semantic\_label\_of\_lane} & Provides boolean values regarding the lane segment, including \textit{is\_intersection} \\ & \textit{turn\_direction}, and \textit{has\_traffic\_control}. \\
 \hline
 \texttt{get\_lane\_ids\_in\_xy\_bbox} & Gets all lane IDs within a Manhattan distance search radius in the $xy$ plane. \\
 \hline
 \texttt{get\_lane\_segment\_predecessor\_ids} & Retrieves all lane IDs with an incoming edge into the query lane segment in the \\ & semantic graph. \\
 \hline
 \texttt{get\_lane\_segment\_successor\_ids} & Retrieves all lane IDs with an outgoing edge from the query lane segment. \\
 \hline
 \texttt{get\_lane\_segment\_adjacent\_ids} & Retrieves all lane segment IDs of that serve as left/right neighbors to the query \\ & lane segment. \\
 \hline
 \texttt{get\_lane\_segment\_centerline} & Retrieves polyline coordinates of query lane segment ID. \\
 \hline
 \texttt{get\_lane\_segment\_polygon} & Hallucinates a lane polygon based around a centerline using avg. lane width. \\
 \hline 
 \texttt{get\_lane\_segments\_containing\_xy} & Uses a ``point-in-polygon'' test to find lane IDs whose hallucinated lane polygons  \\
 & contain this $(x,y)$ query point. \\
 \hline
\end{tabular}
\caption{Example Python functions in the Argoverse map API.}
\label{tab:MapAPIfunctions}
\end{center}
\end{figure*}

\begin{figure}[!t]
    \centering
    \includegraphics[width=\linewidth]{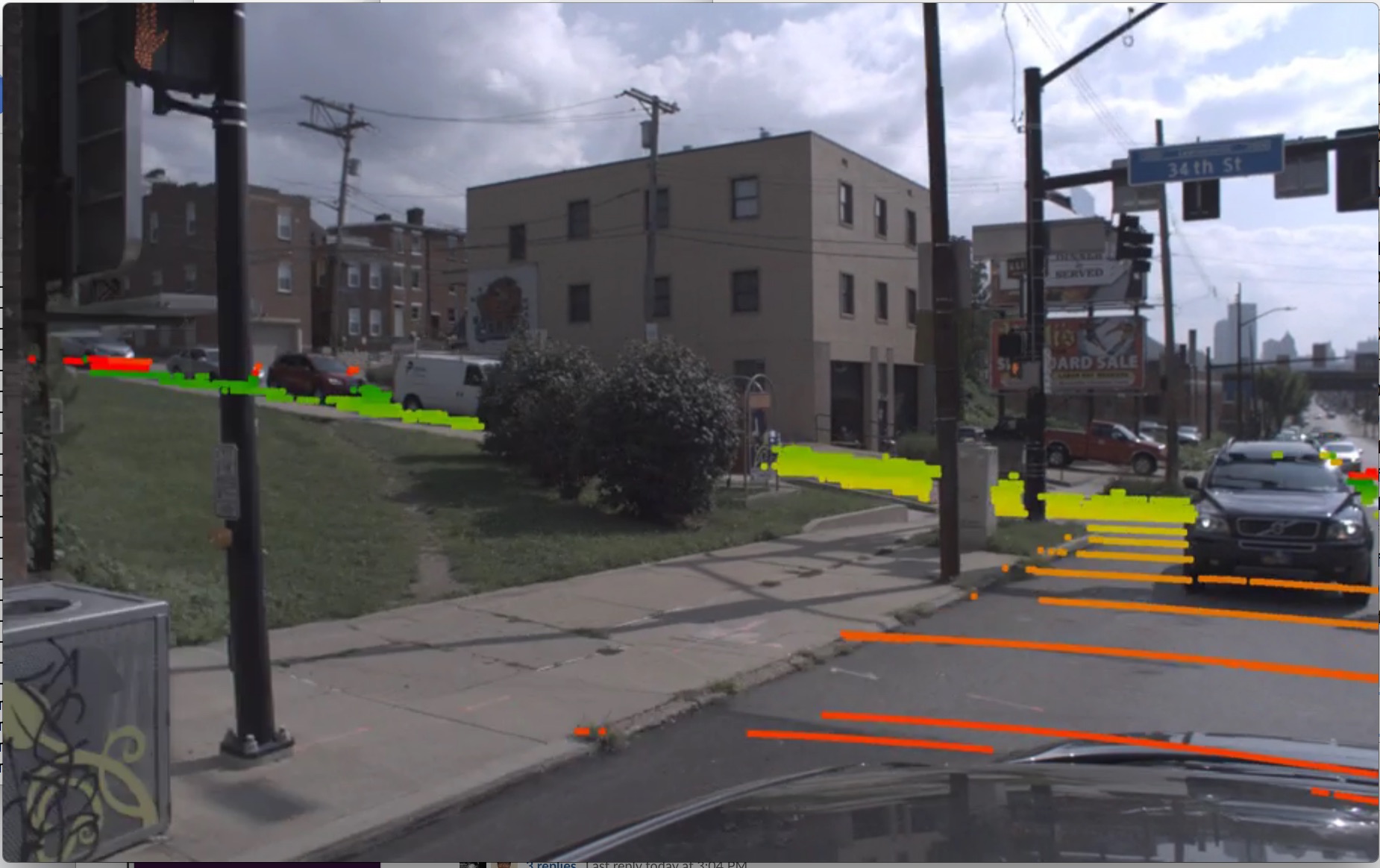}
    \caption{A scene with non-planar ground surface. The colored LiDAR returns have been classified as belonging to the ground, based on the map. Points outside the driveable area are also discarded. This simple distance threshold against a map works well, even on the road to the left which goes steeply uphill.}
    \label{fig:nonplanarground}
\end{figure}

The dataset's rich maps are a novelty for autonomous driving datasets and we aim to make it easy to develop computer vision tools that leverage the map. Figure \ref{tab:MapAPIfunctions} outlines several functions which we hope will make it easier for researchers to access the map. Our API is provided in Python. For example, our API can provide rasterized bird's eye view (BEV) images of the map around the egovehicle, extending up to 100 m in all directions. It can also provide a dense 1 meter resolution grid of the ground surface, especially useful for ground classification when globally planar ground surface assumptions are violated (see Figure \ref{fig:nonplanarground}).

These dense, pixel-level map renderings, similar to visualizations of instance-level or semantic segmentation \cite{Cordts:2016:Cityscapes}, have recently been demonstrated to improve 3D perception and are relatively easy to use as an input to a convolutional network \cite{HDNET:2018:Yang,Casas:2018:IntentNet}.

We provide our vector map data in a modified OpenStreetMap (OSM) format, i.e. consisting of ``Nodes'' (waypoints) composed into ``Ways'' (polylines) so that the community can take advantage of open source mapping tools built to handle OSM formats. The data we provide is richer than existing OSM data which does not contain per-lane or elevation information.

\section{3D Tracking Taxonomy Details}
\label{supp:class-details}

Argoverse 3D Tracking version 1.1 contains 15 categories of objects. Examples of each object type are visualized in Table~\ref{supp:trackingclasses}. Definitions of each of the 15 categories are given below.

\begin{itemize}
    \item \textbf{Animal} A four-legged animal (primarily dogs or cats).
    \item  \textbf{Bicycle} A non-motorized vehicle with 2 wheels that is propelled by human power pushing pedals in a circular motion.
    \item  \textbf{Bicyclist} A person actively riding a bicycle.
    \item  \textbf{Bus} A standard, city bus that makes frequent stops to embark or disembark passengers.
    \item  \textbf{Emergency vehicle} A vehicle with lights and sirens that, when active, gains right-of-way in all situations.
    \item  \textbf{Large vehicle} Motorized vehicles with 4 or more wheels, larger than would fit in a standard garage.
    \item  \textbf{Moped} A motorized vehicle with 2 wheels with an upright riding position with feet together. 
    \item  \textbf{Motorcycle} A motorized vehicle with 2 wheels where the rider straddles the engine.
    \item  \textbf{Motorcyclist} A person actively riding a motorcycle or a moped.
    \item  \textbf{On Road Obstacle} Static obstacles on driveable surface.
    \item  \textbf{Other Mover} Movable objects on the road that don't fall into other categories.
    \item  \textbf{Pedestrian} A person that is not driving or riding in/on a vehicle. 
    \item  \textbf{Stroller} A push-cart with wheels meant to hold a baby or toddler.
    \item  \textbf{Trailer} A non-motorized vehicle towed behind a motorized vehicle. 
    \item  \textbf{Vehicle} Motorized automobile- typically 4-wheels that could fit into a standard, personal garage.
    
\end{itemize}

\begin{table*}[h!]
  \centering
  \begin{tabular}{  c c c c c  }
     \hline
      ANIMAL &  BICYCLE & BUS & EMERGENCY  & LARGE\\
       &  BICYCLIST &  & VEHICLE & VEHICLE \\
      \hline
      \includegraphics[width=3.0cm, height=3.0cm]{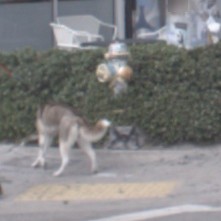} & \includegraphics[width=3.0cm, height=3.0cm]{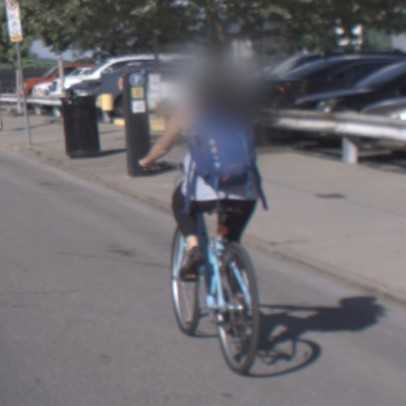} & \includegraphics[width=3.0cm, height=3.0cm]{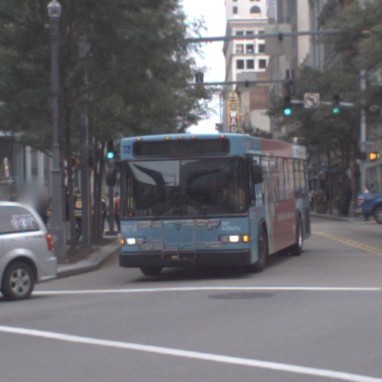} & \includegraphics[width=3.0cm, height=3.0cm]{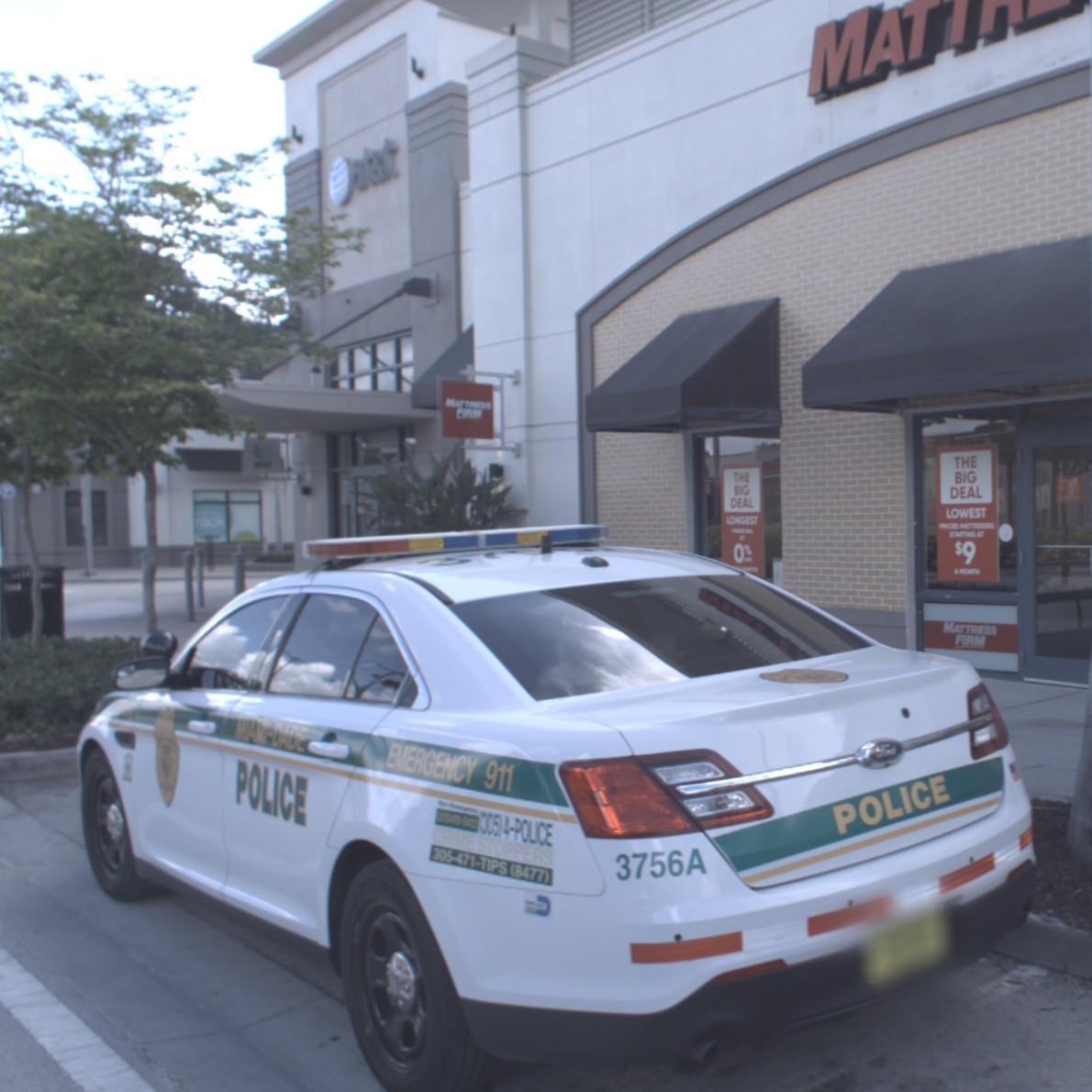} &  \includegraphics[width=3.0cm, height=3.0cm]{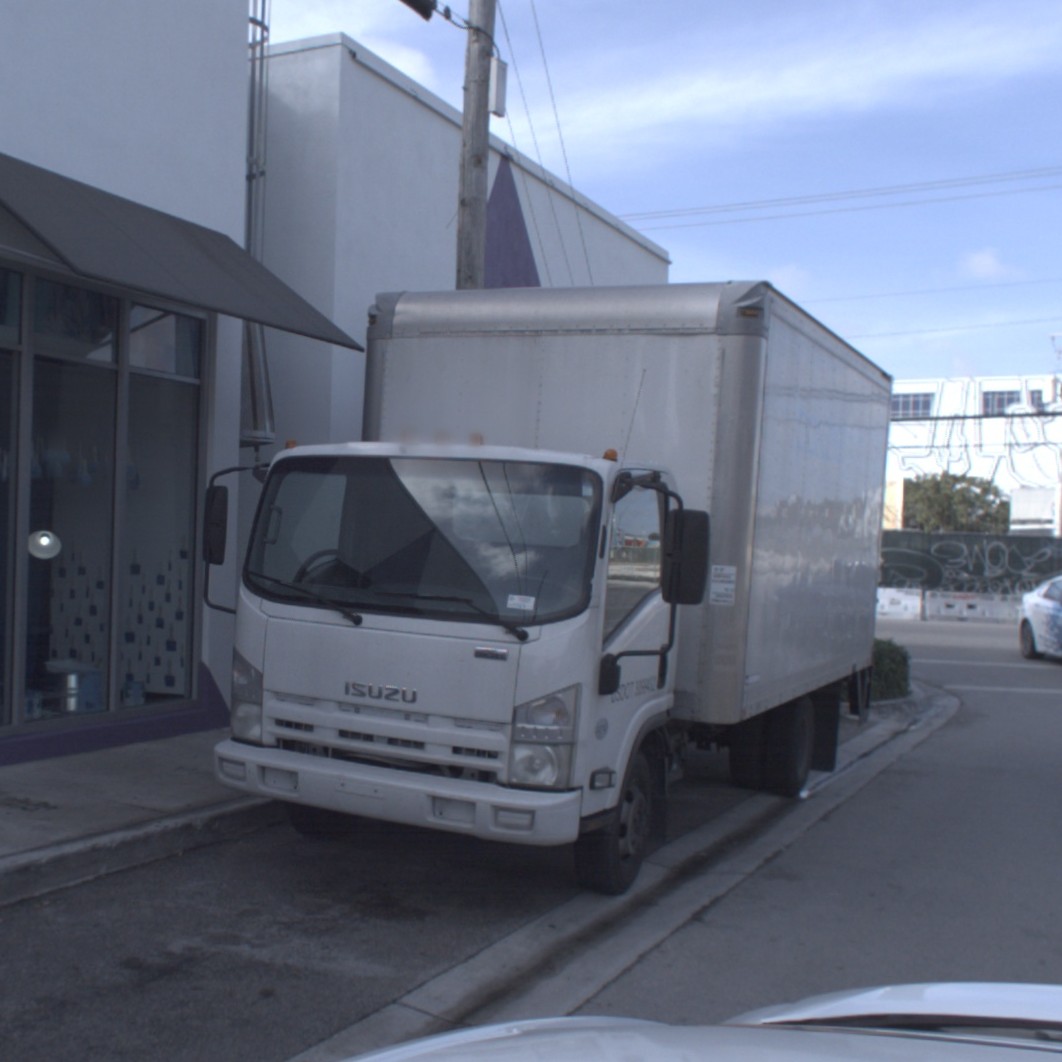} \\
      \hline
        MOPED & MOTORCYCLE   & ON ROAD   & OTHER MOVER &  PEDESTRIAN\\
         & MOTORCYCLIST   & OBSTACLE &  & \\
      \hline
      \includegraphics[width=3.0cm, height=3.0cm]{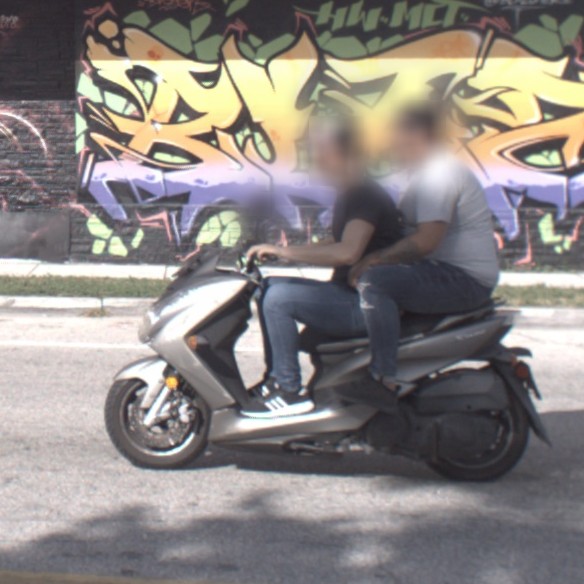} & \includegraphics[width=3.0cm, height=3.0cm]{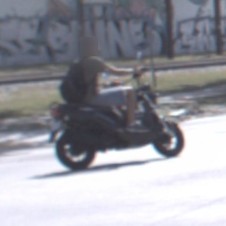} & \includegraphics[width=3.0cm, height=3.0cm]{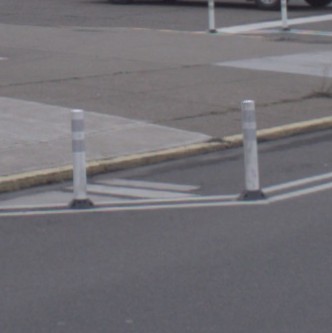} &  \includegraphics[width=3.0cm, height=3.0cm]{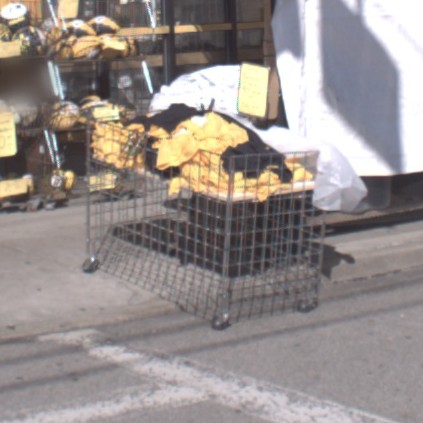} & \includegraphics[width=3.0cm, height=3.0cm]{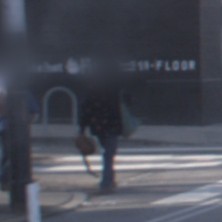}\\
     \hline 
      STROLLER & TRAILER &  VEHICLE  \\
     \hline 
      \includegraphics[width=3.0cm, height=3.0cm]{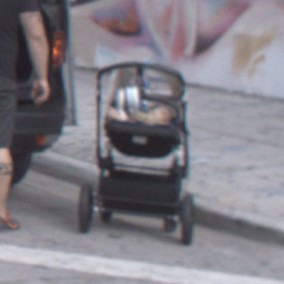} & \includegraphics[width=3.0cm, height=3.0cm]{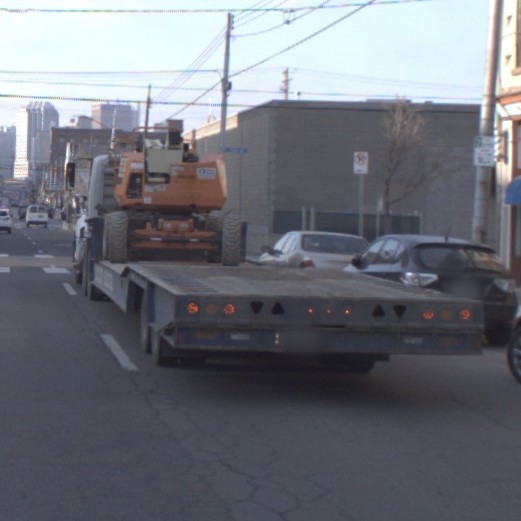} & \includegraphics[width=3.0cm, height=3.0cm]{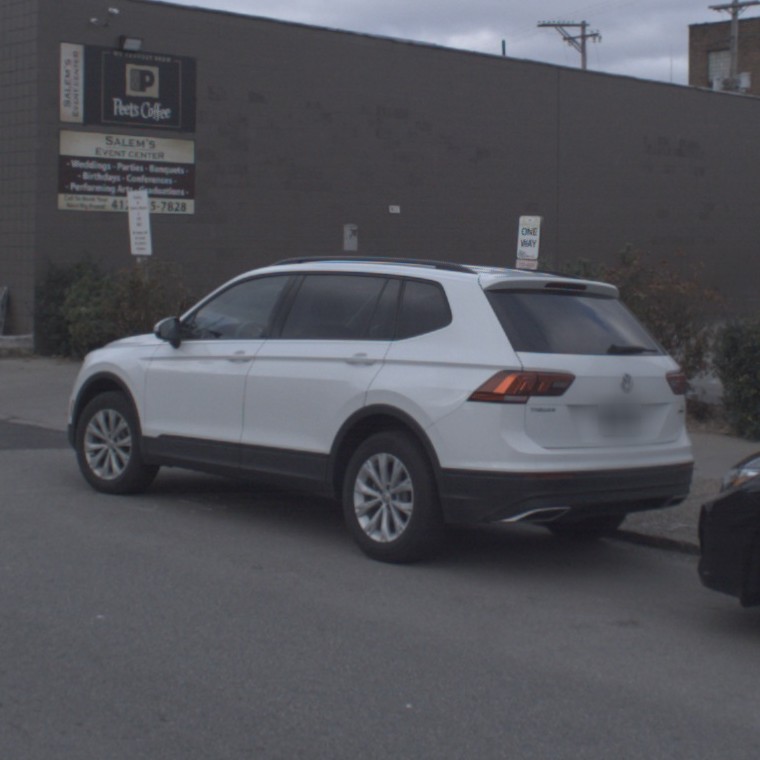} 
  \end{tabular}
  \caption{Argoverse 3D Tracking version 1.1 categories.}\label{supp:trackingclasses}
\end{table*}

\section{Supplemental Details on Motion Forecasting}
\label{supp:traj-mining-details}

In this appendix, we elaborate on some aspects of forecasting data mining and baselines.

\subsection{Motion Forecasting Data Mining Details}

Here we describe our approach for mining data for trajectory forecasting. The scenarios that are challenging for a forecasting task are rare, but with a vector map they are easy to identify. We focus on some specific behavioral scenarios from over 1006 driving hours. For every 5 second sequence, we assign an \textit{interesting} score to every track in that sequence. A high \textit{interesting} score can be attributed to one or more of the following cases wherein the track is: at an intersection with or without traffic control, on a right turn lane, on a left turn lane, changing lanes to a left or right neighbor, having high median velocity, having high variance in velocity and visible for a longer duration. We give more importance to changing lanes and left/right turns because these scenarios are very rare. If there are at least 2 sufficiently important tracks in the sequence, we save the sequence for forecasting experiments. Furthermore, the track which has the maximum \textit{interesting} score and is visible through out the sequence is tagged as the \textit{Agent}. The forecasting task is then to predict the trajectory of this particular track, where all the other tracks in the sequence can be used for learning social context for the \textit{Agent}. There is also a 2.5 second overlap between 2 consecutive sequences. This overlap implies that the same track id can be available in 2 sequences, albeit with different trajectories.

\subsection{2D Curvilinear Centerline Coordinate System}

\begin{figure}
    \centering

    \includegraphics[width=0.75\linewidth]{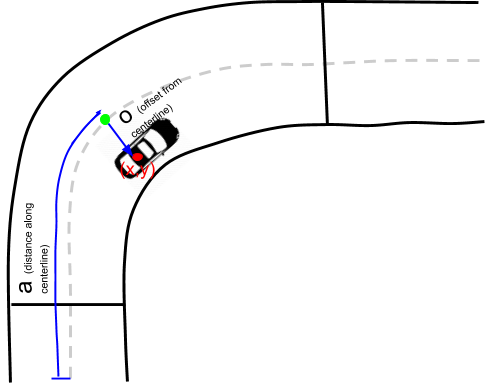}
    \caption{Converting a coordinate $(x,y)$ from the absolute map frame to the 2D curvilinear centerline coordinate frame $(a,o)$ where a vehicle's position is described by a distance $a$ along a centerline  and an offset $o$ from the centerline.}
    \label{fig:map_to_centerline_frame}
\end{figure}

As discussed in Section \ref{forecasting_pipeline}, the baselines that use the map as a prior first transform the trajectory to a 2D Curvilinear Centerline coordinate system. In this section, we provide details about this new coordinate space. The centerline coordinate system has axes tangential and perpendicular to lane centerline. When a trajectory is transformed from the absolute map frame to the centerline frame, it makes the trajectory generalizable across different map locations and orientations. Figure \ref{fig:map_to_centerline_frame} illustrates the transformation.

\section{Supplemental Tracking Details}
\label{supp:trackingdetails}

In this appendix, we describe our tracking pipeline in greater detail.

\subsection{Tracker Implementation Details}

Because of space constraints we provide details of our 3D tracker here instead of in the main manuscript. We do not claim any novelty for this ``baseline'' tracker, but it works reasonably well, especially with map information available (e.g. driveable area, ground height, and lane information). More recent attempts in 3D tracking and detection include the baseline introduced in H3D \cite{patil:h3d:ICRA:2019}, with VoxelNet \cite{Zhou:2018:CVPR:VoxelNet} detection and an Unscented Kalman Filter \cite{julier:1995:new, julier:2000:new} for tracking. The NuScenes dataset \cite{holger:nuscenes:2019} uses PointPillars \cite{lang:pointpillars:cvpr2019} for a 3D bounding box detection baseline. ApolloCar3D \cite{ApolloCar3D:song:cvpr:2019} implements 2D to 3D car pose estimation baselines based on 3D-RCNN\cite{3DRCNN_CVPR18} and DeepMANTA \cite{DeepMANTA_CVPR17}. However, we do not compare our baseline tracker against these methods.

Our tracker tracks the position and velocity of surrounding vehicles from LiDAR data. The tracking pipeline has the following stages:

\textbf{1. Segmentation and Detection.}
In order to segment a point cloud into distinct object instances, we exploit the complementary nature of our two sensor modalities. First, we use geometrically cluster the remaining 3D LiDAR point cloud into separate objects according to density, using DBSCAN \cite{Ester96adensity-based}. Then we use Mask R-CNN \cite{he:2017:maskrcnn} to obtain object masks in pixel space and discard any LiDAR clusters whose image projection does not fall within a mask.  

Others have proposed compensating for point cloud undersegmentation and oversegmentation scenarios by conditioning on the data association and then jointly track and perform probabilistic segmentation \cite{Held:RSS:2016}. We can avoid many such segmentation failures with the high precision of our Mask R-CNN network\footnote{We use a public implementation available at \url{https://github.com/facebookresearch/maskrcnn-benchmark}.}. We also alleviate the need for an object's full, pixel-colored 3D shape
during tracking, as others have suggested \cite{Held:RSS:2014, held:2013:precision}. We prefer density-based clustering to connected components clustering in a 2D occupancy grid \cite{himmelsbach:08:lidar-based3d, Levinson:2011:towardsfully}  because the latter approach discards information along the z-axis, often rendering the method brittle.


To help focus our attention to areas that are important for a self driving car, we only consider points within the driveable region defined by the map. We also perform ground removal using either the ground height map or plane-fitting. 
    
While segmentation provides us a set of points belonging to an object, we need to determine if this is an object of interest that we want to track. Unlike in image space, objects in a 3D have consistent sizes. We apply heuristics that enforce the shape and volume of a typical car and thereby identify vehicle objects to be tracked. We estimate the center of an object by fitting a smallest enclosing circle over the segment points. 
    
\textbf{2. Association.} We utilize the Hungarian algorithm to obtain globally optimal assignment of previous tracks to currently detected segments where the cost of assignment is based on spatial distance. Typically, tracks are simply assigned to their nearest neighbor in the next frame.

\textbf{3. Tracking.} We use ICP (Iterative Closest Point) from the Point Cloud Library \cite{PCL:2011} to estimate a relative transformation between corresponding point segments for each track. Then we apply a Kalman Filter (KF) \cite{held:2013:precision} with ICP results as the measurement and a static motion model (or constant velocity motion model, depending on the environment) to estimate vehicle poses for each tracked vehicle. We assign a fixed size bounding box for each tracked object. The KF state is comprised of both the 6 dof pose and velocity.


\subsection{Tracking Evaluation Metrics}

We use standard evaluation metrics commonly used for multiple object trackers (MOT) \cite{MOT16,Bernardin2008EvaluatingMO}. The MOT metric relies on centroid distance as distance measure.

\begin{itemize}
    \item {\bf{MOTA}(Multi-Object Tracking Accuracy)}:
        \begin{equation}
            MOTA = 100*(1-\frac{\sum_t{FN_t+FP_t+ID_{sw}}}{\sum_{t}GT})
        \end{equation}
        where $FN_t$, $FP_t$, $ID_{sw}$, $GT$ denote the number of false negatives, false positives, number of ID switches, and ground truth objects. We report MOTA as percentages.
        
    \item {\bf{MOTP}(Multi-Object Tracking Precision)}:
        \begin{equation}
            MOTP=\frac{\sum_{i,t}D^i_t}{\sum_t{C_t}}
        \end{equation}    
        where $C_t$ denotes the number of matches, and $D^i_t$ denotes the distance of matches.  

    \item {\bf{IDF1} (F1 score)}:
        \begin{equation}  
            IDF1 = 2\frac{precision * recall}{precision+recall}
        \end{equation}    
        Where $recall$ is the number of true positives over number of total ground truth labels. $precision$ is the number of true positives over sum of true positives and false positives.


    \item {\bf{MT} (Mostly Tracked)}: Ratio of trajectories tracked more than 80\% of its lifetime.
    \item {\bf{ML} (Mostly Lost)}: Ratio of trajectories tracked for less than 20\% of object lifetime, over the entire object lifetime.
    \item {\bf{FP} (False Positive)}: Total number of false positives.
    
    \item {\bf{FN} (False Negative)}: Total number of false negatives.

    \item {\bf{IDsw} (ID Switch)}: Number of identified ID switches.
    
    \item {\bf{Frag} (Fragmentation)}: Total number of switches from "tracked" to "not tracked".
    
\end{itemize}

\subsection{True Positive Thresholding Discussion}
\begin{figure}[ht]
\centering
\subfloat[Fixed inter-centroid distance of $\sqrt{2}$ m.]{
  \includegraphics[width=0.7\linewidth]{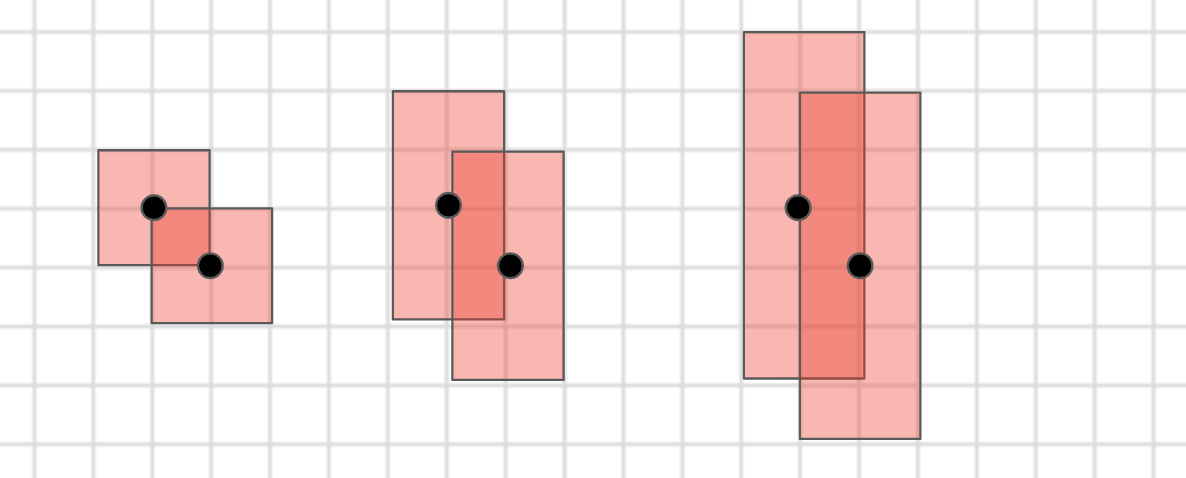}
}
\hspace{0mm}
\subfloat[Fixed 2D intersection area of 2 $m^2$.]{
  \includegraphics[width=0.7\linewidth]{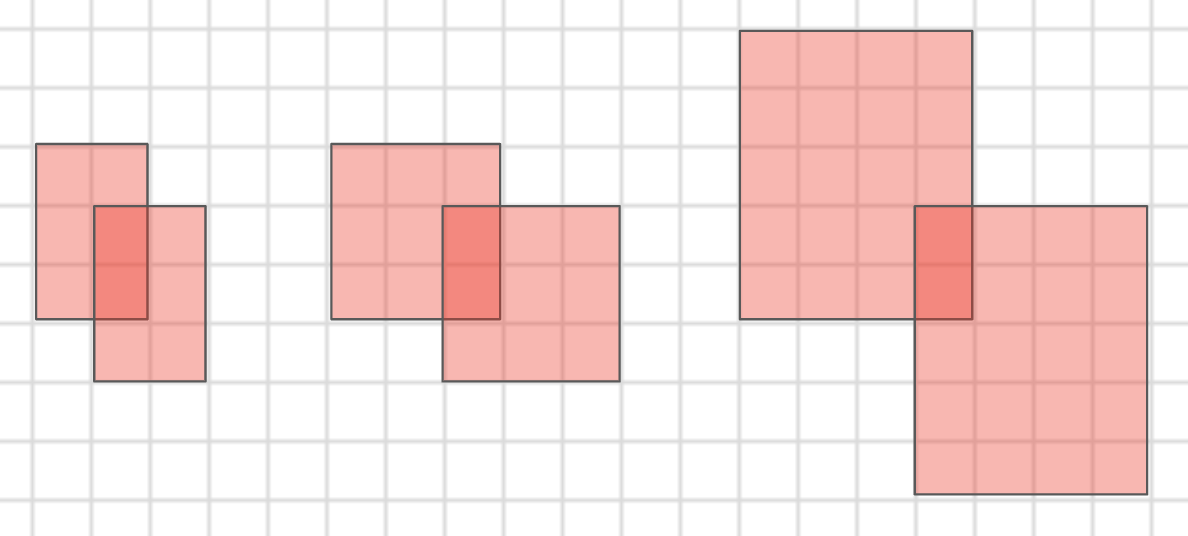}
}
 \caption{ We compare thresholding true positives (TP) and false positives (FP) of matched cuboid shapes using inter-centroid distance (above) versus using 2D/3D IoU (below). Above: fixed inter-centroid distance, from left to right: IoU values of 0.143, 0.231, 0.263. 
 Below: fixed intersection area, from left to right, IoU values of 0.2, 0.125, 0.053. 
 }\label{fig:centroidvsiou}

\end{figure}

Intersection-over-Union (IoU) is designed as a scale invariant metric, meaning that doubling the size and relative overlap of two boxes will not change its value. However, we counter that 3D tracking evaluation should not be performed in a strictly scale invariant manner. \emph{Absolute} error matters, especially in 3D. In 2D tasks (e.g. object detection) we operate on \emph{pixels} which could be any real world size, whereas in 3D we have absolute lengths. When using IoU as a TP/FP threshold, small objects tend to be penalized unfairly. For example, for pairs of bounding boxes with the same distance between centroids (e.g. a pedestrian that is tracked with 0.5 m error vs a car that is tracked with 0.5 m error), the larger objects will have higher IoU (see Figure \ref{fig:centroidvsiou}). To track pedestrians with the same IoU as buses requires orders of magnitude more positional precision.

In the LiDAR domain, these problems are exaggerated because the sampling density can be quite low, especially for distant objects. In 2D object detection, we rarely try to find objects that are 3 pixels in size, but small, distant objects frequently have 3 LiDAR returns and thus accurate determination of their spatial extent is difficult. Still, the centroids of such objects can be estimated. For these reasons, we use the absolute distance between centroids as the basis for classifying correct vs. incorrect associations between tracked and ground truth objects.

\begin{figure}
\centering
\subfloat[Argoverse LiDAR]{
  \includegraphics[width=0.4\linewidth]{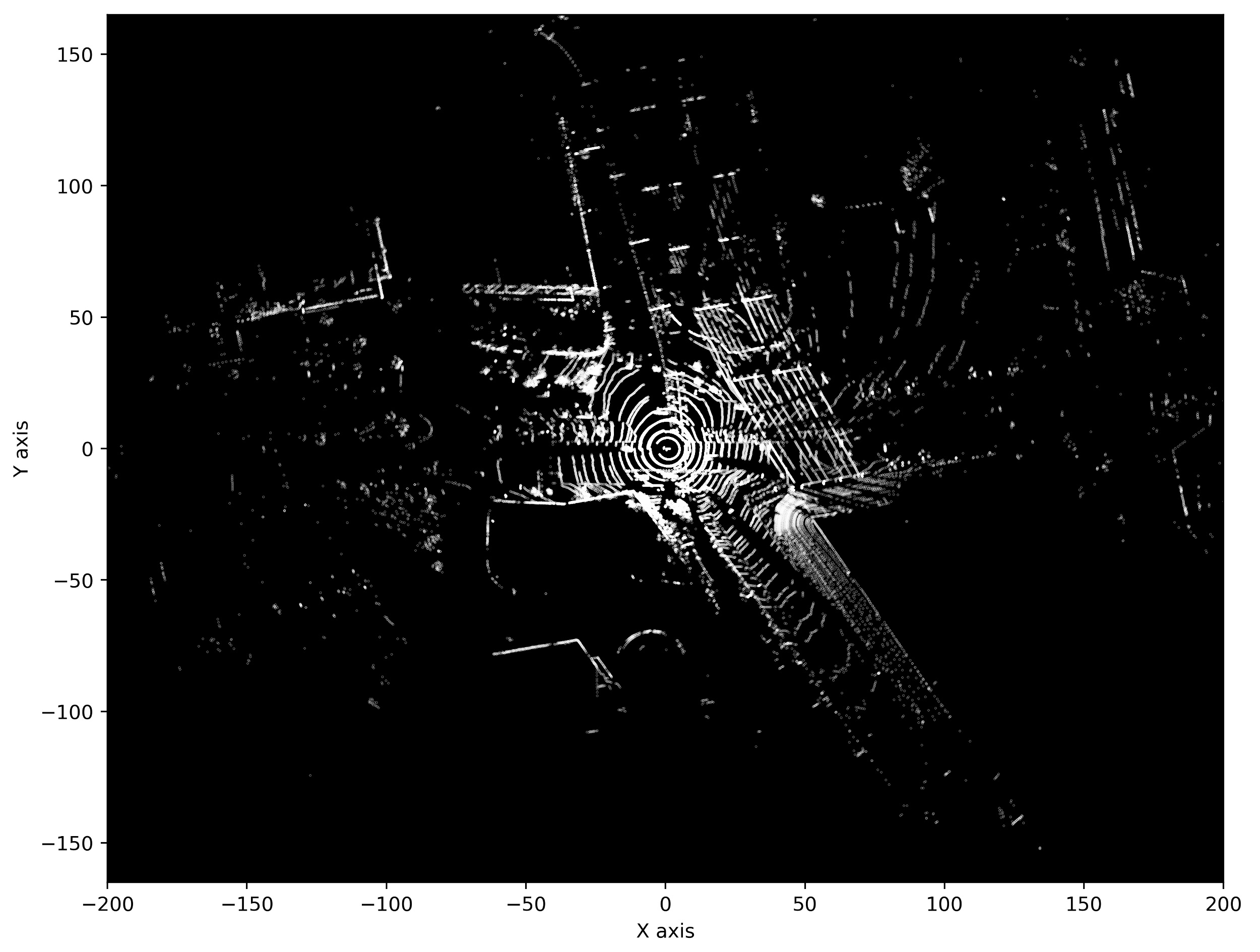}
}
\subfloat[Argoverse LiDAR]{
   \includegraphics[width=0.4\linewidth]{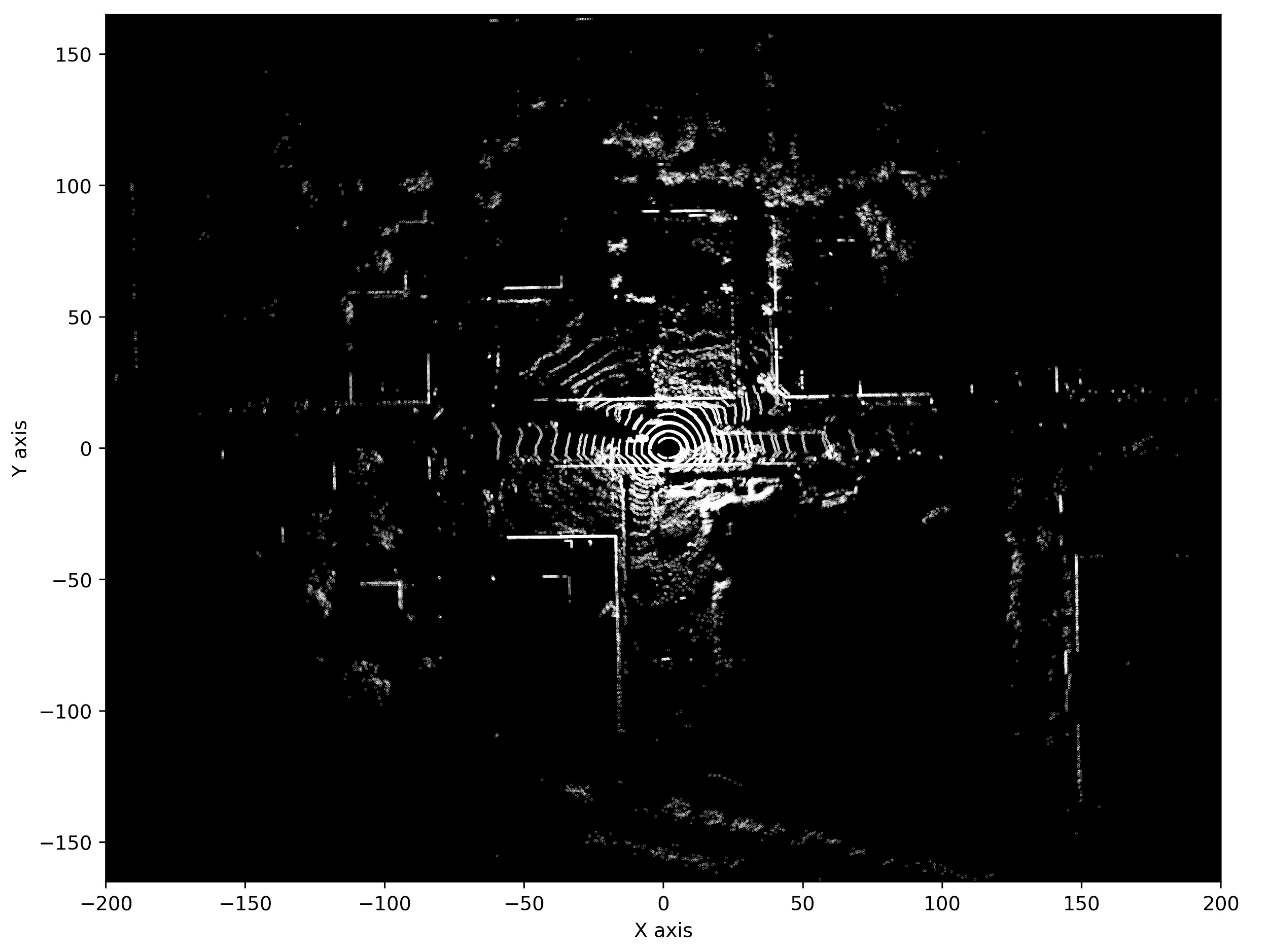}
}
\hspace{0mm}
\subfloat[KITTI LiDAR]{
  \includegraphics[width=0.4\linewidth]{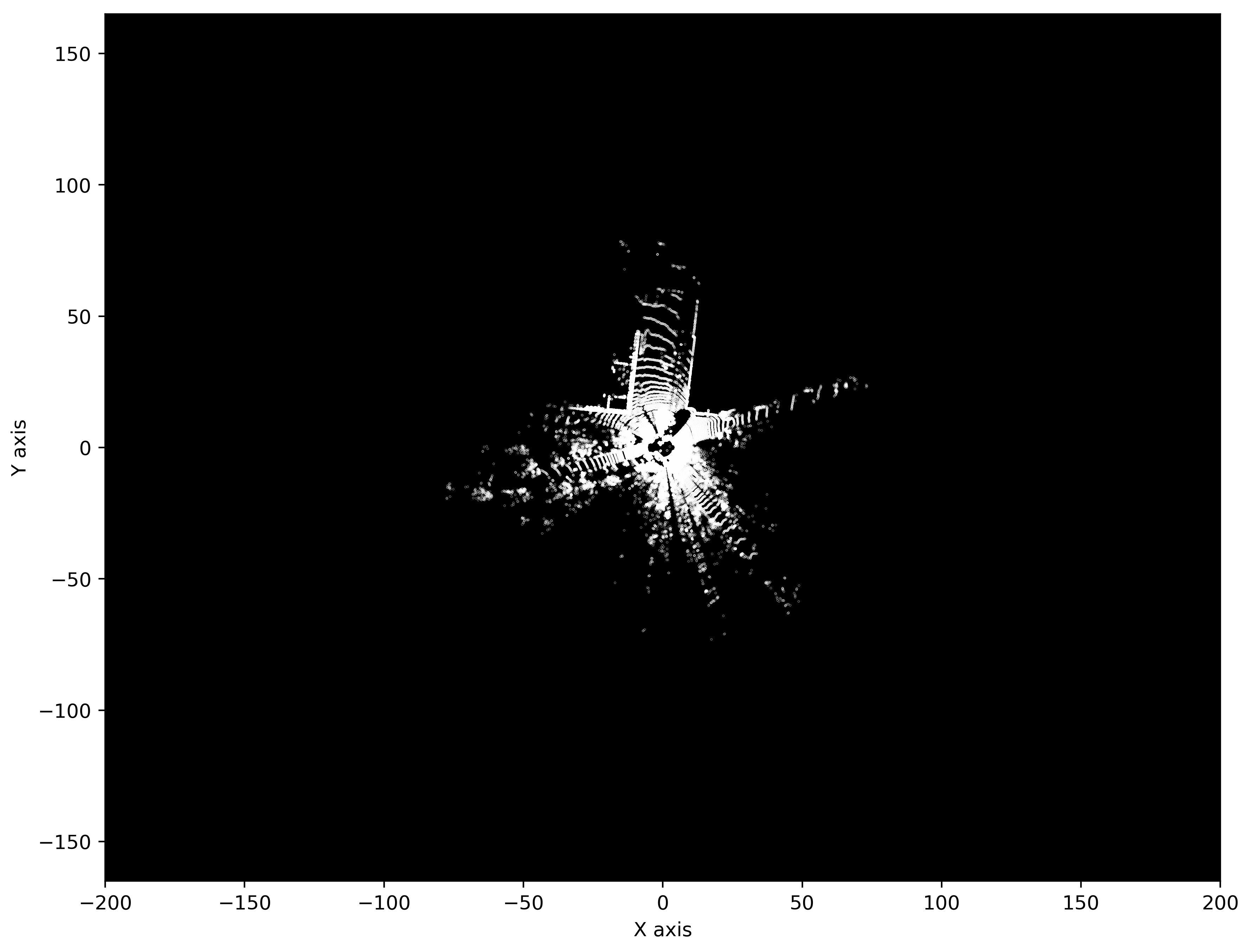}
}
\subfloat[nuScenes LiDAR]{
   \includegraphics[width=0.4\linewidth]{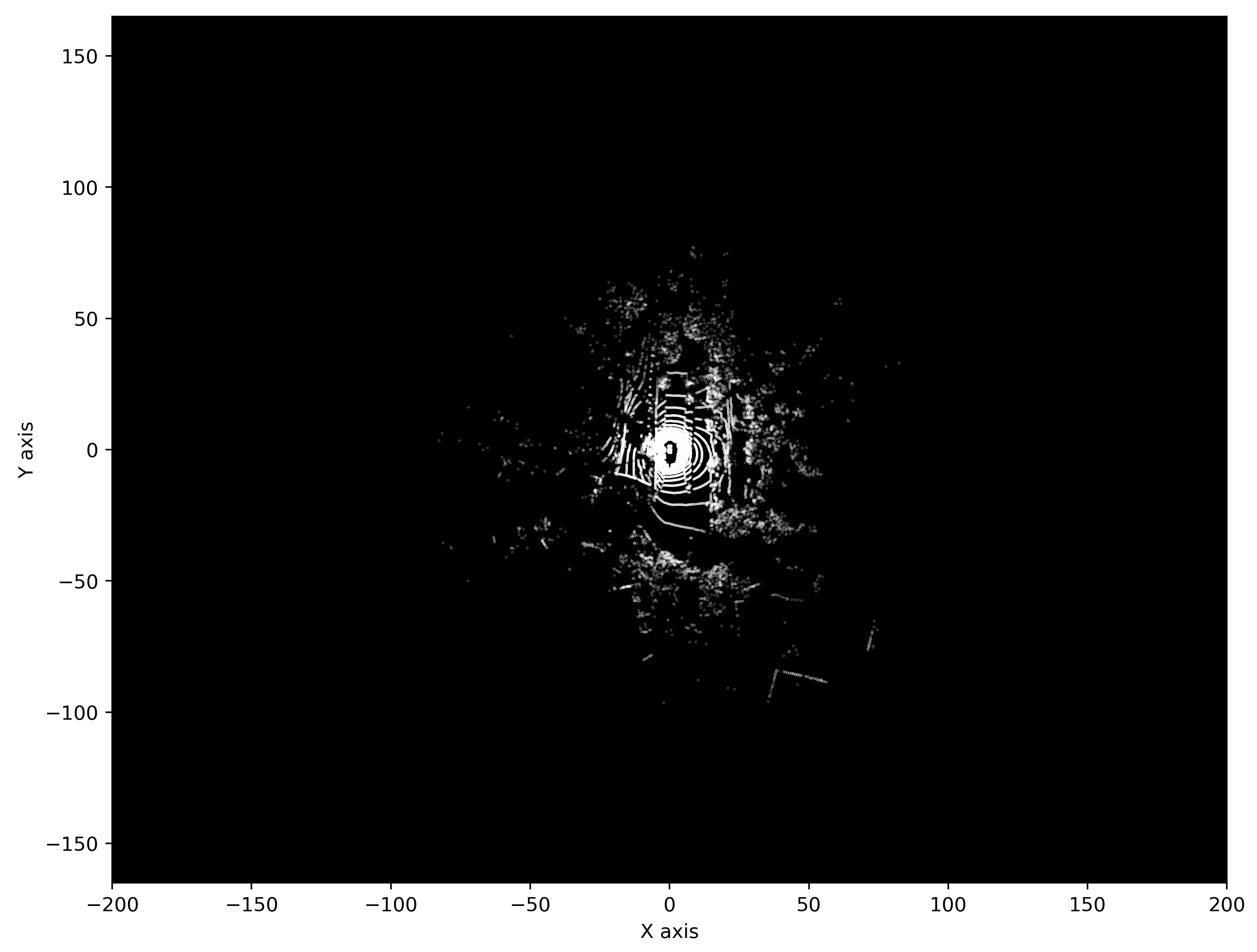}
}
 \caption{Above: Sample LiDAR sweeps in the ego-vehicle frame, with marked $x$ and $y$ axes, with $x\in [-200,200]$ and $y\in [-160,160]$ for all plots. The Argoverse LiDAR has up to twice the range of the sensors used to collect the KITTI or nuScenes datasets, allowing us to observe more objects in each scene.}\label{fig:LiDAR_range}

\end{figure}

\end{document}